\documentclass[10pt,conference,a4paper]{IEEEtran}
\usepackage{cite}

\ifCLASSINFOpdf
   \usepackage[pdftex]{graphicx}
\else
   \usepackage[dvips]{graphicx}
\fi
\usepackage{amsmath}
\usepackage{algorithmic}

\usepackage{array}

\ifCLASSOPTIONcompsoc
 \usepackage[caption=false,font=normalsize,labelfont=sf,textfont=sf]{subfig}
\else
 \usepackage[caption=false,font=footnotesize]{subfig}
\fi
\usepackage{fixltx2e}

\usepackage{stfloats}
\usepackage{url}

\DeclareMathOperator*{\argmax}{argmax}

\DeclareMathOperator*{\sign}{sign}

\begin{document}
%
\title{Time Series Data Augmentation \\ for Neural Networks by Time Warping \\ with a Discriminative Teacher}

\author{\IEEEauthorblockN{Brian Kenji Iwana}
\IEEEauthorblockA{Department of Advanced Information Technology\\
Kyushu Univeristy, Fukuoka, Japan\\
Email: brian@human.ait.kyushu-u.ac.jp}
\and
\IEEEauthorblockN{Seiichi Uchida}
\IEEEauthorblockA{Department of Advanced Information Technology\\
Kyushu Univeristy, Fukuoka, Japan\\
Email: uchida@ait.kyushu-u.ac.jp}}


%


\maketitle

\begin{abstract}
Neural networks have become a powerful tool in pattern recognition and part of their success is due to generalization from using large datasets. However, unlike other domains, time series classification datasets are often small. In order to address this problem, we propose a novel time series data augmentation called guided warping. While many data augmentation methods are based on random transformations, guided warping exploits the element alignment properties of Dynamic Time Warping (DTW) and shapeDTW, a high-level DTW method based on shape descriptors, to deterministically warp sample patterns. In this way, the time series are mixed by warping the features of a sample pattern to match the time steps of a reference pattern. Furthermore, we introduce a discriminative teacher in order to serve as a directed reference for the guided warping. We evaluate the method on all 85 datasets in the 2015 UCR Time Series Archive with a deep convolutional neural network (CNN) and a recurrent neural network (RNN). The code with an easy to use implementation can be found at \url{https://github.com/uchidalab/time_series_augmentation}.
\end{abstract}


%

\section{Introduction}

In recent times, deep neural networks have become commonplace and continue to set the state-of-the-art benchmarks across many domains, such as natural scene object classification~\cite{NIPS2019_9035}, machine translation~\cite{Edunov_2018}, graph classification~\cite{verma2019learning}, and more. 
Part of the recent successes is due to the increase in data availability and the advancement of hardware to support it~\cite{Schmidhuber_2015}. 
In fact, it is well-known that increasing the amount of data helps with generalization and, in turn, the accuracy of many machine learning models~\cite{brain1999on,Banko_2001,Torralba_2008}. 

However, unlike the image domain, time series datasets tend to be tiny in comparison. 
For example, one of the most used sources of time series classification datasets, the University of California Riverside~(UCR) Time Series Archive~\cite{UCRArchive}, contains 85 time series datasets but only 10 have more than 1,000 training samples and the largest, ElectricDevices, only has 8,926. 
By comparison, the popular image datasets, ImageNet Large Scale Visual Recognition Challenge~(ILSVRC)~\cite{ILSVRC15}, MNIST~\cite{Lecun_1998}, and CIFAR~\cite{krizhevsky2009learning}, have 1.2 million, 60,000, and 50,000 training patterns respectively. 
Thus, in order to use the full potential of modern machine learning methods, there is a need for time series classification data.

One solution to tackle this problem is the use of data augmentation. 
Specifically, data augmentation is a common data-space solution to increase the generalization ability of machine learning models. 
It does this by increasing the size of the training dataset using synthetic patterns. 
Data augmentation has shown to be an effective model-independent method of reducing overfitting and expanding the decision boundary modeled by the data~\cite{Shorten_2019}. 

\begin{figure}[!t]
\centering
\subfloat[Proposed Random Guided Warping]{\includegraphics[width=0.97\columnwidth,trim=0cm 0.8cm 0cm 0cm,clip]{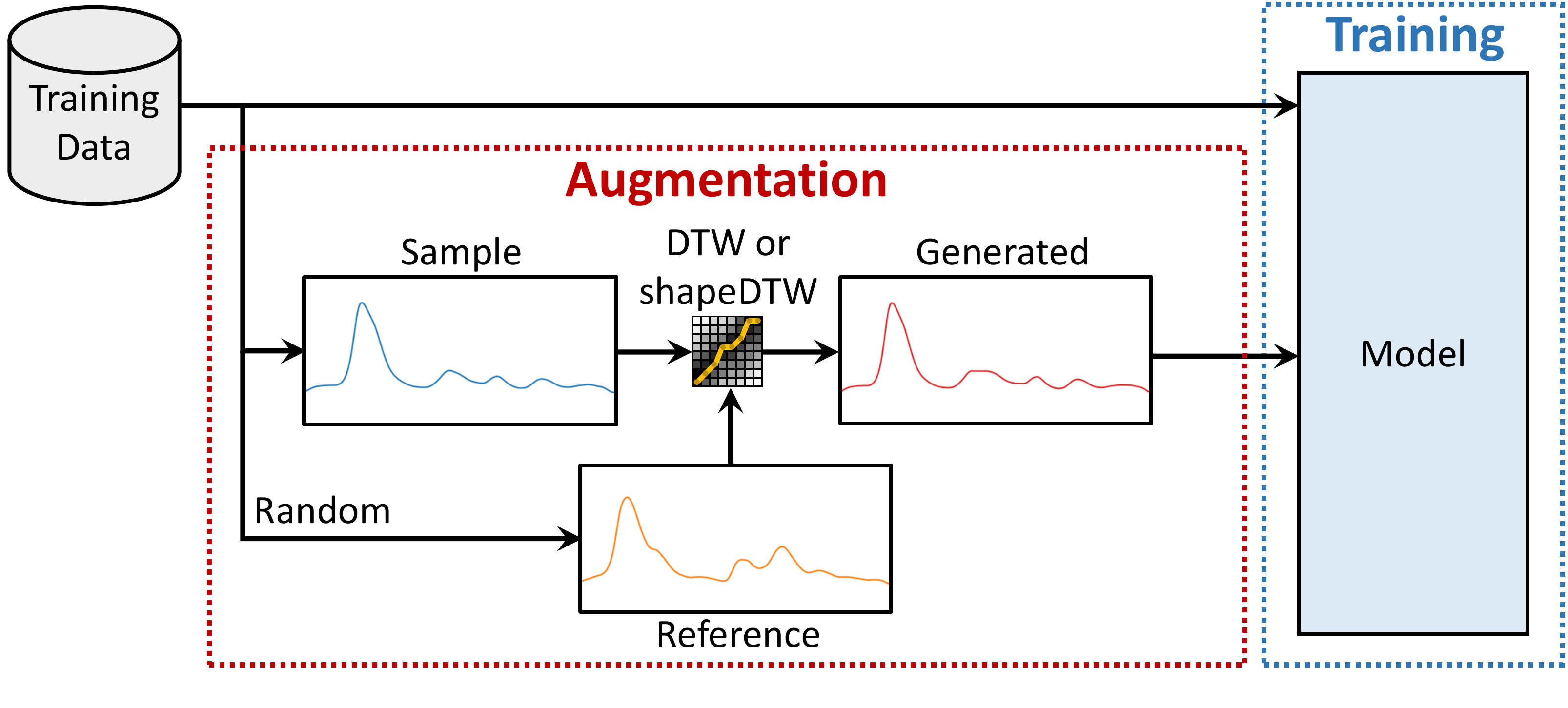}}

\vspace{-3mm}
\subfloat[Proposed Discriminative Guided Warping]{\includegraphics[width=0.97\columnwidth,trim=0cm 0cm 0cm 0cm,clip]{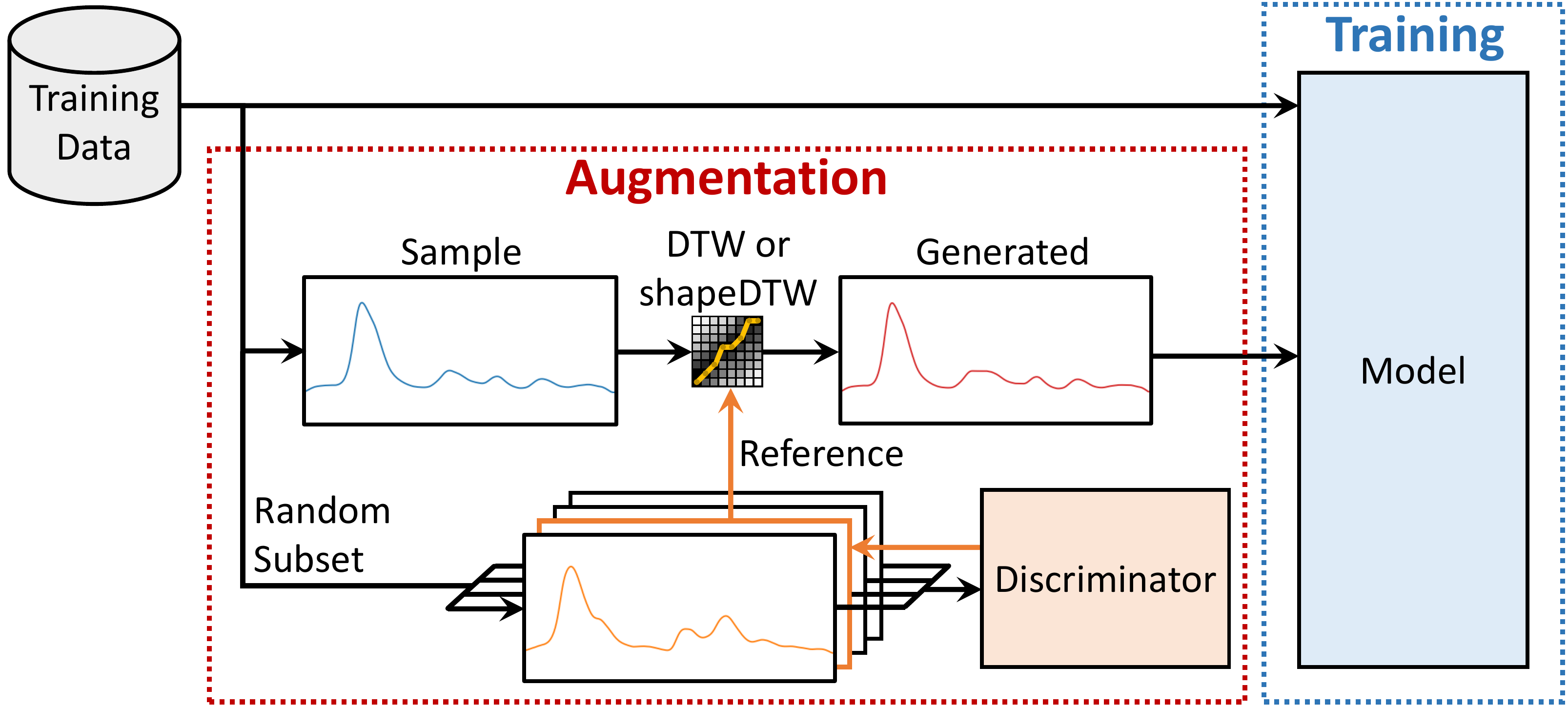}}
\caption{Workflow of the proposed method. For data augmentation, in (a), two patterns are selected and DTW is used to warp a sample by the reference. In (b), the sample is warped by a reference selected by a discriminator using a small subset of samples. The generated patterns are used together with the original training data to train a model.
\vspace{-3mm}
}
\label{fig:diagram}
\end{figure}

Data augmentation for images is a well-explored field, especially in combination with neural networks. 
It has almost become a standard practice for image classification. 
For example, many of the popular Convolutional Neural Networks~(CNN) based architectures used a form of data augmentation in their training, such as the original proposals of AlexNet~\cite{krizhevsky2009learning}, Very Deep Convolutional Networks~(VGG)~\cite{simonyan2014very}, and Residual Networks~(ResNet)~\cite{He_2016}.
Comparatively, there are fewer established time series data augmentation methods and fewer standard data augmentation practices in time series classification~\cite{wen2020time}. 
Most of the methods that exist are just time series adaptations inspired by image recognition. 
These methods generally rely on simple transformations, such as jittering (adding noise), scaling, rotation, etc.

However, time series have different properties than images and these methods might not be applicable to all time series. 
In addition, while some time series specific data augmentation methods exist, such as magnitude warping~\cite{Um_2017} and time warping~\cite{Um_2017,Rashid_2019}, they are still random transformations that carry assumptions about the patterns in the underlying dataset. 


In order to tackle the problem of time series data augmentation, we propose the use of a new pattern mixing based augmentation, called \textit{guided warping}. 
Guided warping combines the idea of time warping~\cite{Um_2017} with pattern mixing~\cite{Takahashi_2016}. 
Specifically, to augment the training dataset with new samples, we warp the features of a \textit{sample} to the time step relations of a \textit{reference}. 
To align the features between the two time series, Dynamic Time Warping~(DTW)~\cite{sakoe1978dynamic} is used. 
While typically DTW is used as a distance measure, it can be used for its ability to align similar features while maintaining the temporal properties~\cite{Petitjean_2011,Iwana_2019}. 
Furthermore, we demonstrate that the dynamic warping can be further improved using shapeDTW~\cite{Zhao_2018} due to a smoother alignment by using high-level shape descriptors instead of element-wise alignment. 

Furthermore, there is a question on how to choose the reference time series. 
Existing time series pattern mixing based data augmentation methods ten to either select the mixed patterns at random~\cite{Takahashi_2016,Forestier_2017,Kamycki_2019} or using a medoid~\cite{Forestier_2017}. 
We show that these are not necessarily the best strategies and propose a novel method of using a discriminator for selection, as shown in Fig~\ref{fig:diagram}. 
The reference pattern selected by the discriminator is referred to as a discriminative teacher and it is determined by finding the sample within a bootstrap set with the maximal distance between the patterns of the same class and patterns of a different class.  
For this work, we use a simple nearest centroid classifier on a small batch of random samples. 
Using a discriminative teacher rather than a random teacher, allows the guided warping to directly choose patterns that might aid the classifier. 

The contributions are as follows:
\begin{itemize}
    \item We demonstrate that using a DTW based warping is effective at producing samples for data augmentation.
    \item We show that the use of shapeDTW in the proposed guided warping can help maintain the original features of the sample time series and generate more suitable samples.
    \item We propose the use of a discriminative teacher time series instead of a random reference to be used as the basis of the warping. 
    \item We do a thorough evaluation on all 85 datasets of the 2015 UCR Time Series Archive~\cite{UCRArchive} using nine established time series data augmentation methods and the four proposed methods. The data augmentation methods are evaluated using a temporal 1D VGG~\cite{simonyan2014very} and a Long Short-Term Memory~(LSTM)~\cite{Hochreiter_1997} network.
\end{itemize}

\section{Related Work}
\label{sec:related}

\begin{figure}[!t]
\centering
\subfloat[Original]{\includegraphics[width=0.32\columnwidth,trim=0.4cm 0.4cm 0.3cm 0.3cm,clip]{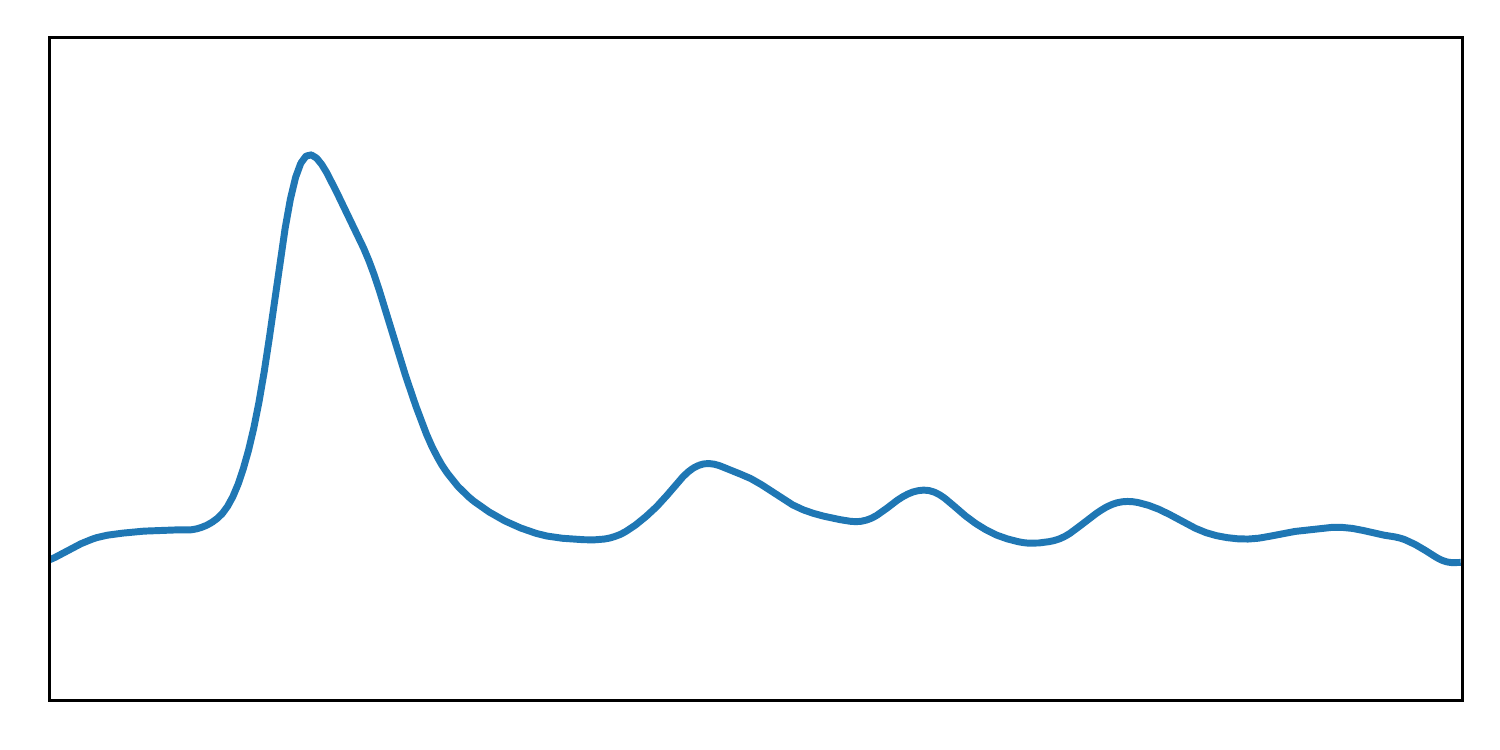}}~
\subfloat[Jittering]{\includegraphics[width=0.32\columnwidth,trim=0.4cm 0.4cm 0.3cm 0.3cm,clip]{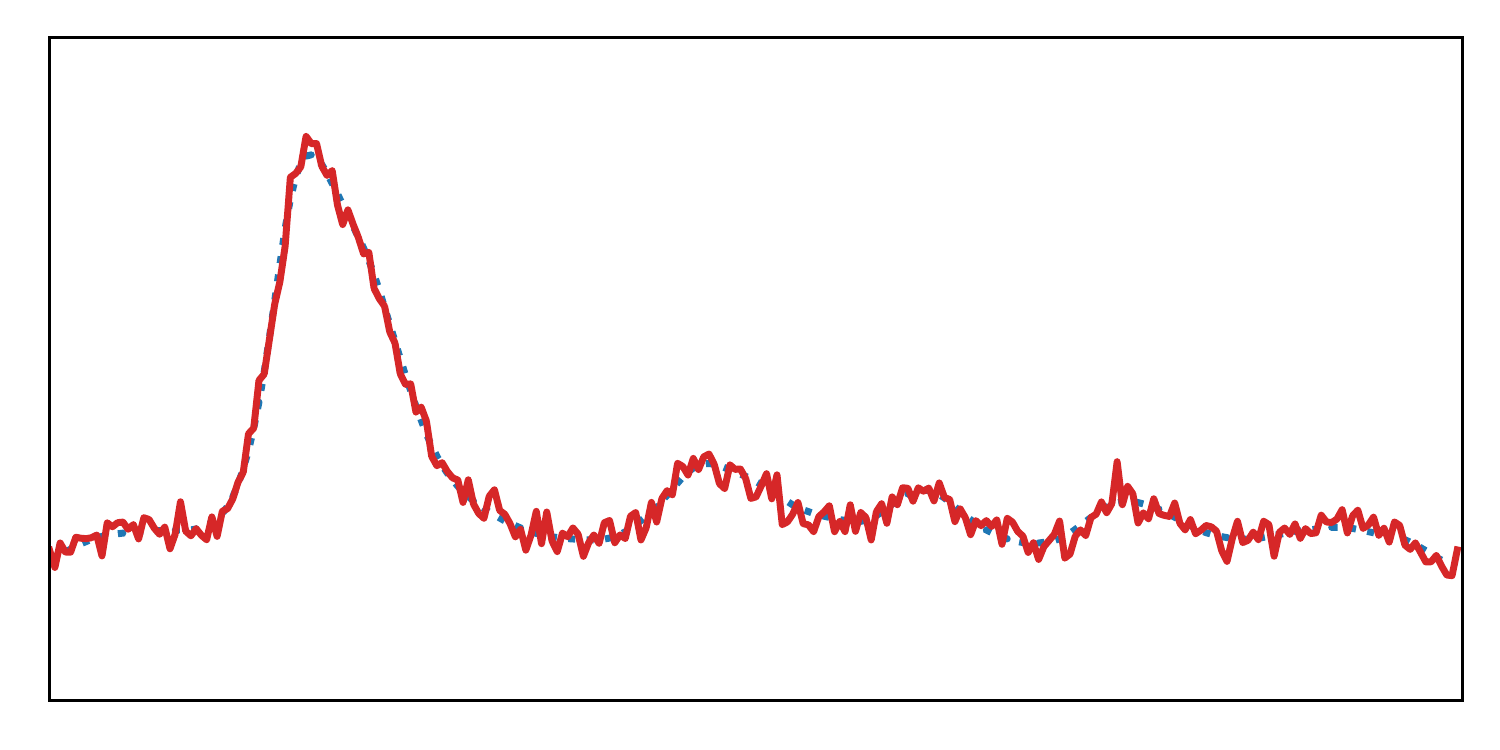}}~
\subfloat[Scaling]{\includegraphics[width=0.32\columnwidth,trim=0.4cm 0.4cm 0.3cm 0.3cm,clip]{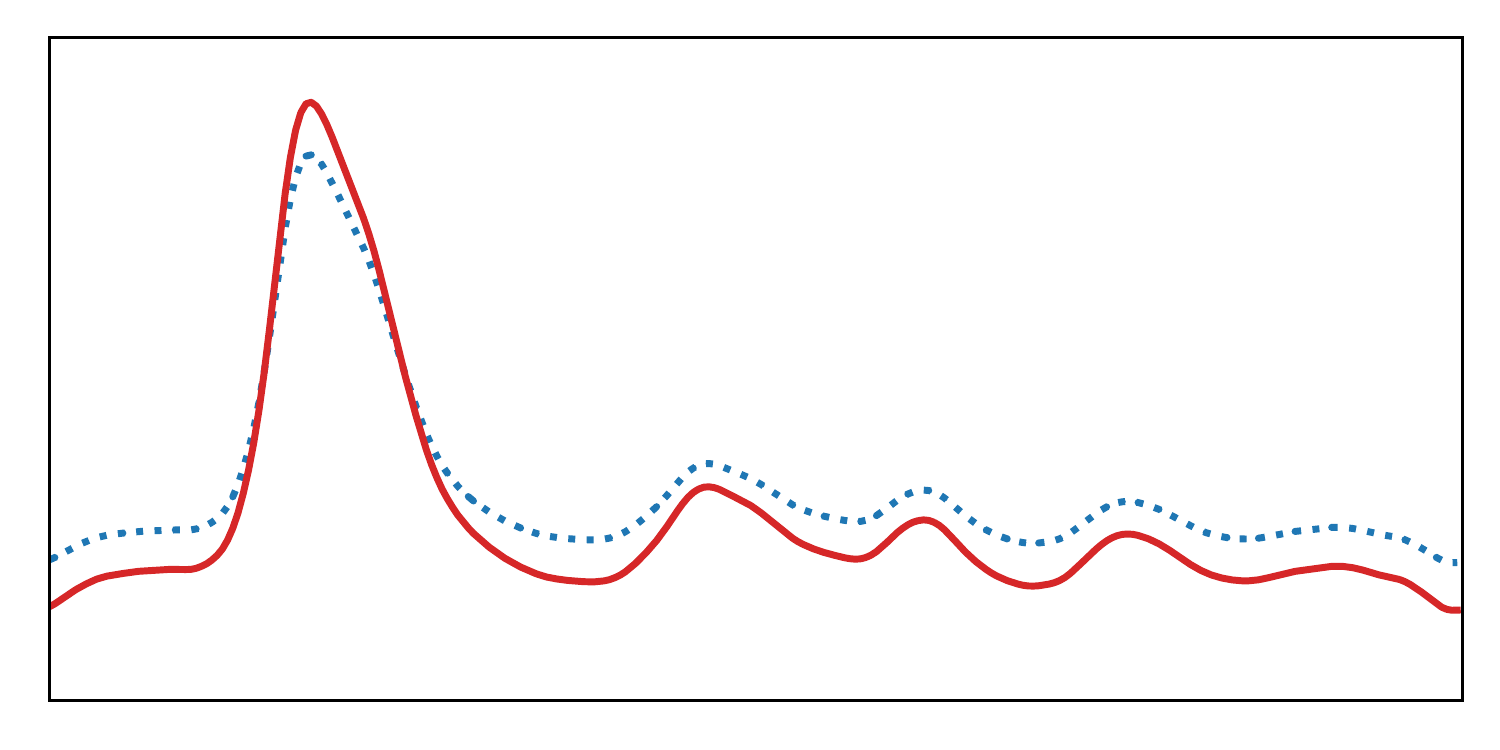}}

\subfloat[Magnitude Warping]{\includegraphics[width=0.32\columnwidth,trim=0.4cm 0.4cm 0.3cm 0.3cm,clip]{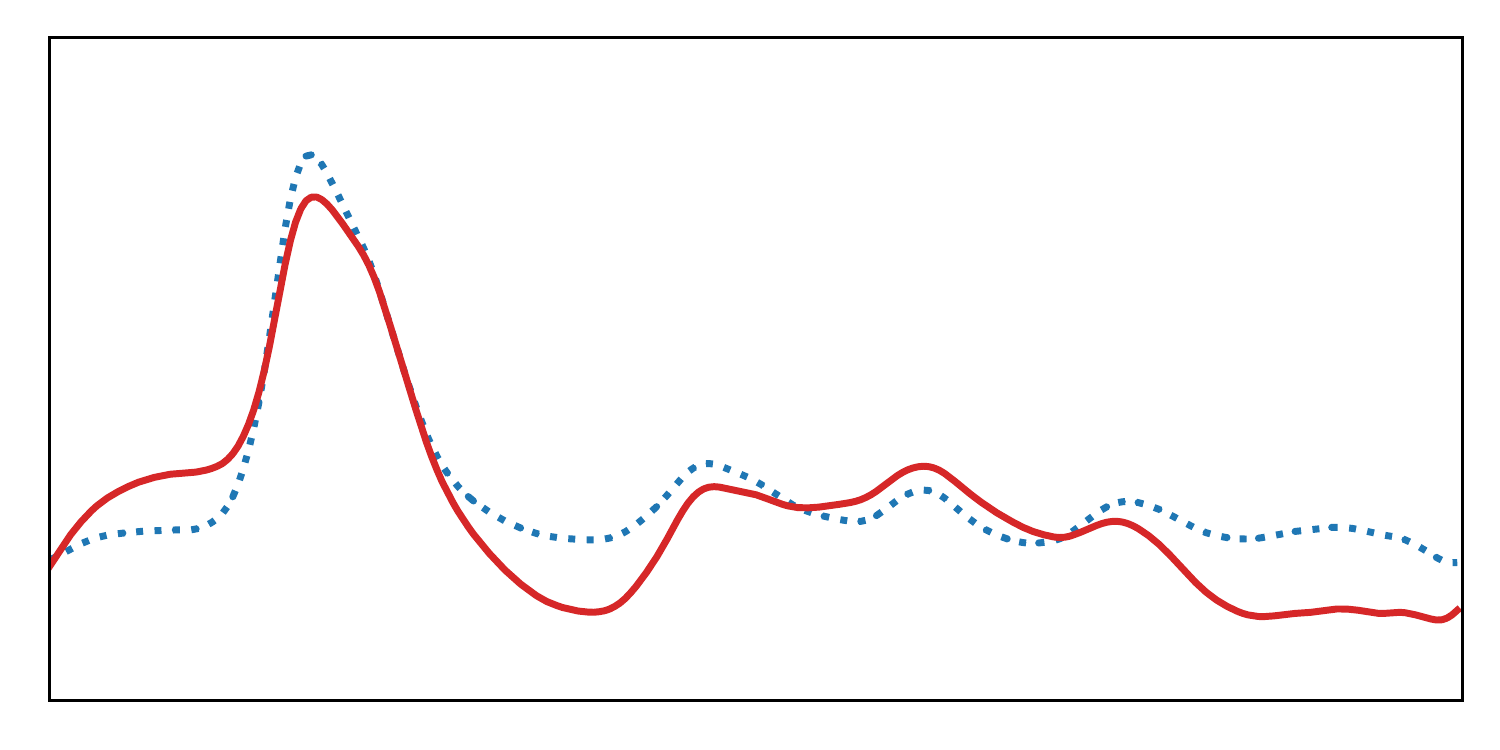}}~
\subfloat[Rotation]{\includegraphics[width=0.32\columnwidth,trim=0.4cm 0.4cm 0.3cm 0.3cm,clip]{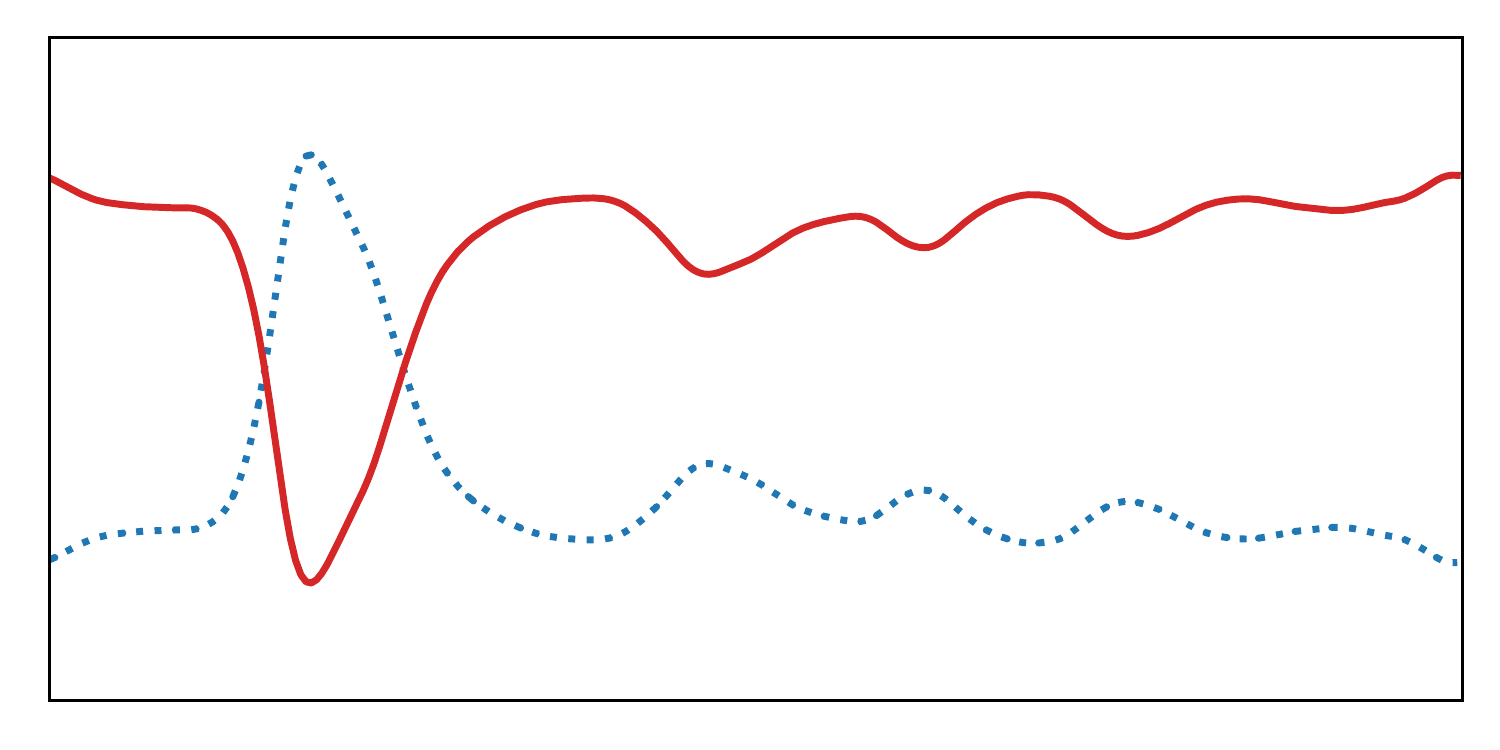}}~
\subfloat[Permutation]{\includegraphics[width=0.32\columnwidth,trim=0.4cm 0.4cm 0.3cm 0.3cm,clip]{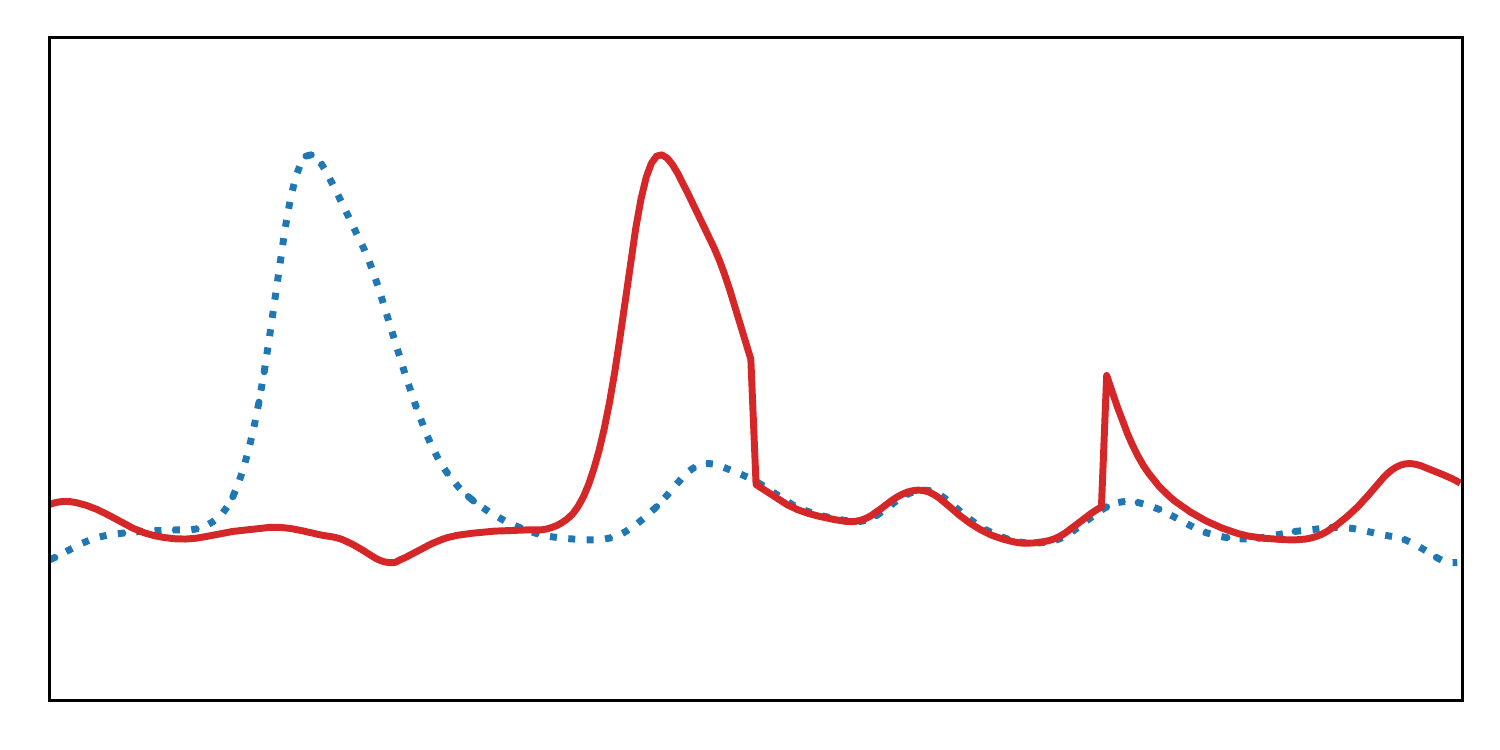}}

\subfloat[Window Slice]{\includegraphics[width=0.32\columnwidth,trim=0.4cm 0.4cm 0.3cm 0.3cm,clip]{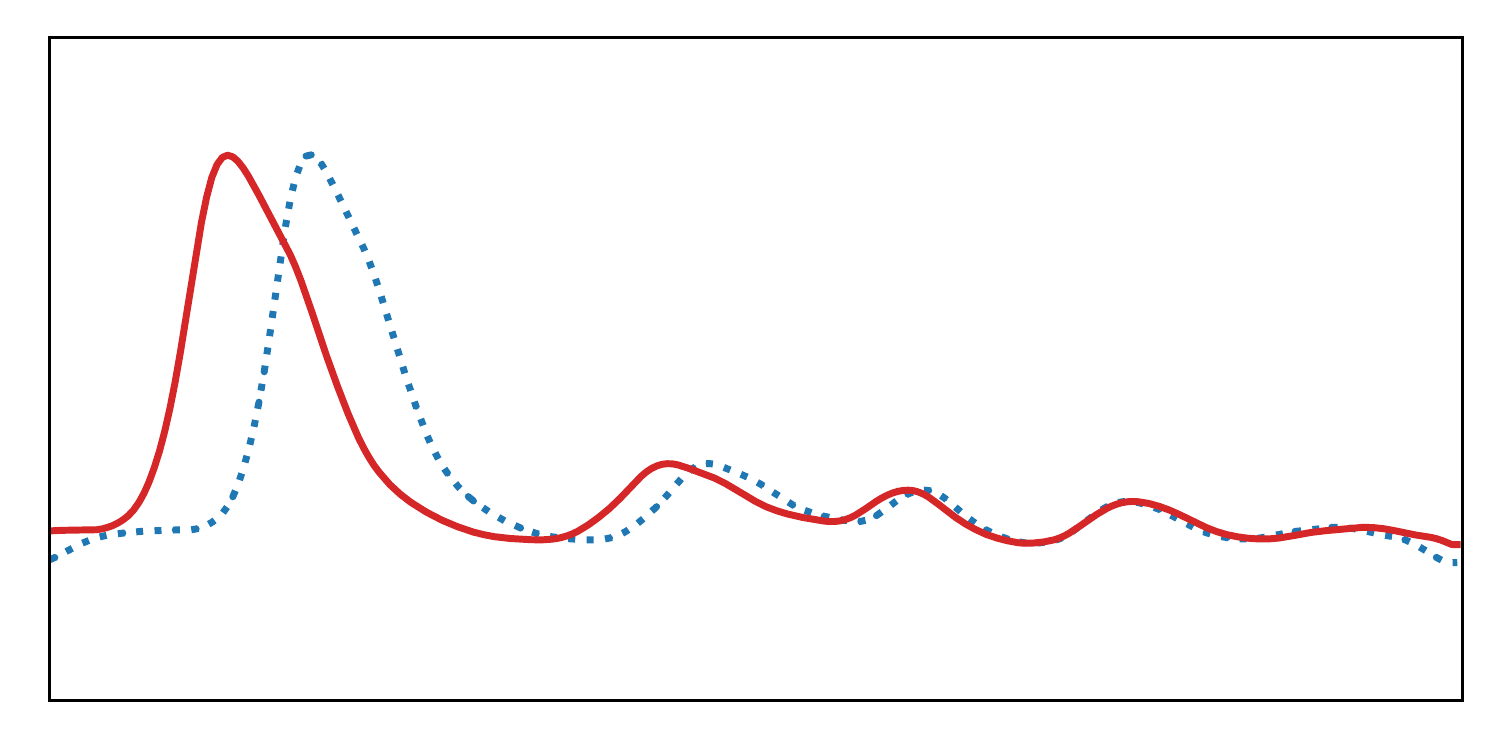}}~
\subfloat[Time Warping]{\includegraphics[width=0.32\columnwidth,trim=0.4cm 0.4cm 0.3cm 0.3cm,clip]{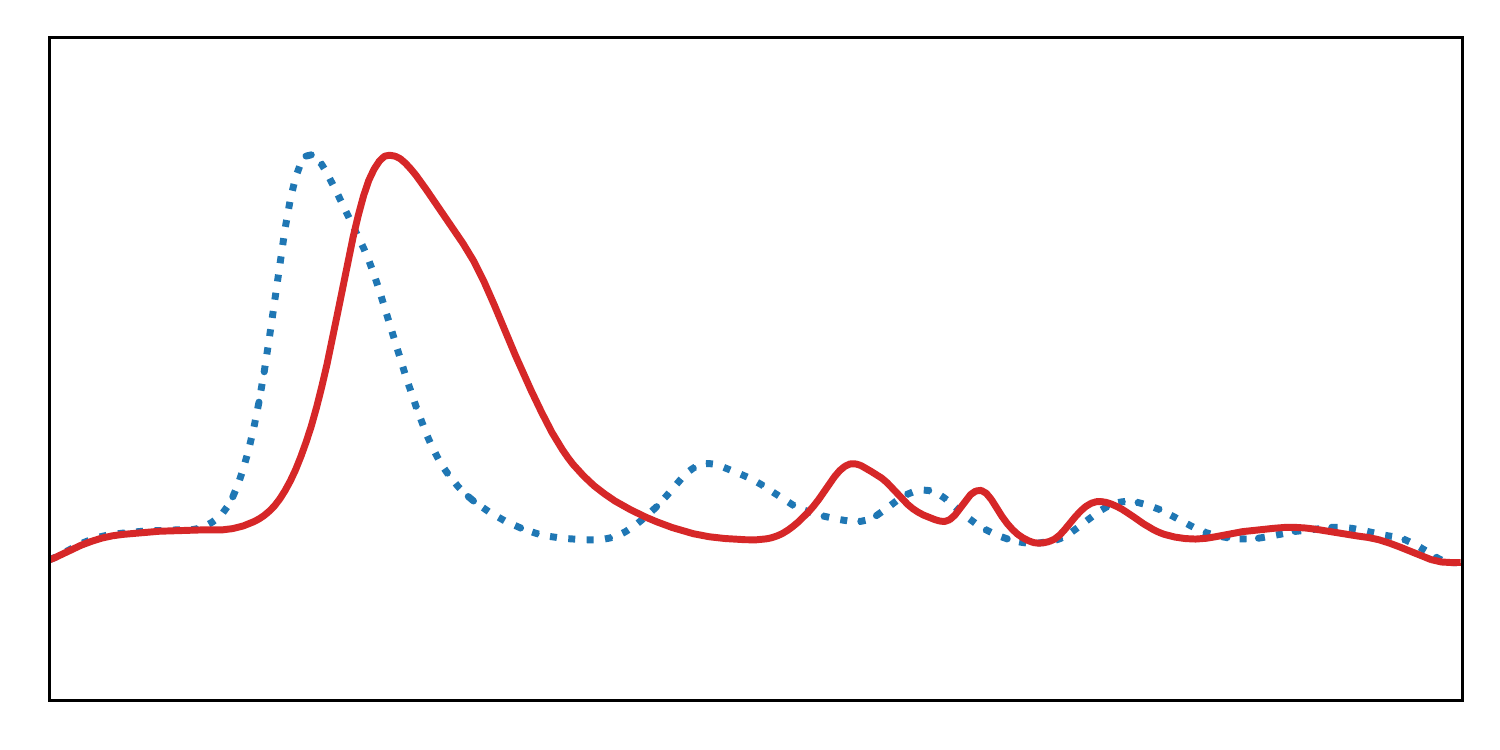}}~
\subfloat[Window Warping]{\includegraphics[width=0.32\columnwidth,trim=0.4cm 0.4cm 0.3cm 0.3cm,clip]{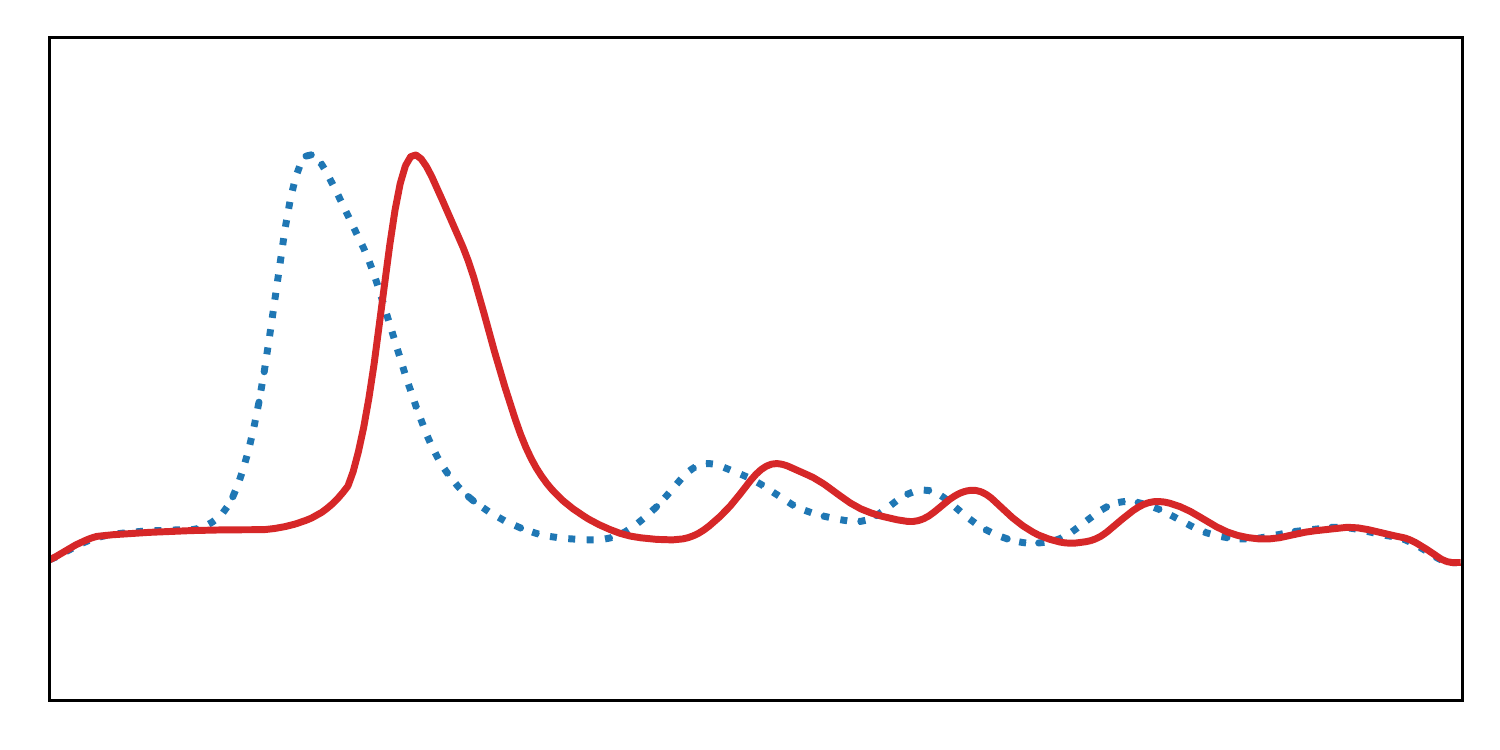}}
\caption{A sample pattern from the 50words dataset in the UCR Time Series Archive. The blue line is the original time series and the red lines are transformed time series that are used as data augmentation.
\vspace{-3mm}
}
\label{fig:examples}
\end{figure}

Most instances of time series data augmentation are random transformations.
Some variations include \textit{jittering} (noise addition), \textit{rotation} (flipping for univariate; rotation for multivariate), \textit{slicing} (cropping), \textit{permutation} (rearranging slices), \textit{scaling} (pattern-wise magnitude change),  \textit{magnitude warping} (smooth element-wise magnitude change), \textit{time warping} (time step deformation), and \textit{frequency warping} (frequency deformation). 
Examples of these are shown in Fig.~\ref{fig:examples}. 
These random transformation based methods have been used for a wide variety of time series.

There are many examples of using random transformations for different applications. 
For example, Um et al.~\cite{Um_2017} used a variety of augmentations, such as jittering, scaling, rotation, slicing, permutation, magnitude warping, and time warping, to improve wearable sensor data for deep temporal CNNs. 
Rashid and Louis~\cite{Rashid_2019} recently used jittering, scaling, rotation, and time warping with LSTMs for construction equipment activity recognition.
Frequency warping, in particular, is often used in acoustic recognition~\cite{jaitly2013vocal,Cui_2014}. 
In addition, there have been improvements on the random transformations. 
An example of this is \textit{window warping}~\cite{le2016data}, which is a version of time warping that expands or contracts random windows of the data by fixed amounts.

Another approach is \textit{pattern mixing}. 
In pattern mixing, instead of adding random transformations, multiple samples of the same class are mixed. 
In one example of pattern mixing, Takahashi et al.~\cite{Takahashi_2016} adds together random segments of intra-class sounds at different ratios. 
However, adding two sequences together might result in out of phase overlapping or in the case of non-periodic time series, malformed patterns. 
Therefore, Forestier et al.~\cite{Forestier_2017} utilized DTW Barycentric Averaging~(DBA)~\cite{Petitjean_2011} to generate patterns. 
Specifically, DBA is an iterative method of using DTW to align time series to find an average pattern with the features preserved. 
Forestier et al. proposed taking small subsets of the data and averaged them using a weighted DBA~(wDBA). 
Other pattern mixing methods include using a randomly weighted DBA~\cite{IsmailFawaz2018}, averaging patterns with sub-optimal time warping~\cite{Kamycki_2019}, and stochastic feature mapping~\cite{Cui_2014}.

Finally, there are also many miscellaneous methods of generating data such as trained generative models~\cite{nikolaidis2019augmenting} and handcrafted mathematical models~\cite{Wendling_2000}. 
In particular, Generative Adversarial Networks (GAN)~\cite{goodfellow2014generative} are a popular method of generating time series data, as they have shown to be useful for data augmentation~\cite{nikolaidis2019augmenting}. 
However, the problem with these methods is that either they require domain specific properties (e.g. mathematical models) or they require external training (e.g. GANs).

The difference between the proposed method and these methods is that we attempt to address the data augmentation problem with as few assumptions as possible. 
The problem with many of the random transformations is that not all transformations are applicable to every dataset. 
For example, something simple as jittering carries the assumption that it is typical for the dataset to have noise. Adding jittering to an ECG dataset seems to fit, however, adding jittering to a dataset with only smooth shape outlines (such as the 50words dataset shown in Fig.~\ref{fig:examples}) does not. 
In the 50words dataset, the heights of cursive words are mapped to a time series and the patterns created from jittering (Fig.~\ref{fig:examples}~(b)) and rotation (Fig.~\ref{fig:examples}~(e)) become unnatural. 
Pattern mixing augmentation can overcome these issues, but the previously proposed pattern mixing methods also have faults. 
For example, DBA is an effective method of averaging time series, but in wDBA averaging similar patterns might not help in increasing the distribution of patterns for better generalization. 

\section{Guided Warping}
\label{sec:generation}

In this section, we will define DTW and propose the use of it as a method to time warp time series and how it can be used for data augmentation.

\subsection{Dynamic Time Warping}
\label{sec:dtw}

DTW~\cite{sakoe1978dynamic} is a classic, yet, effective method of determining an optimized distance measure for time series. 
Consider two time series $\mathbf{r}=r_1,\dots,r_i,\dots,r_I$ and $\mathbf{s}=s_1,\dots,s_j,\dots,s_J$ with sequence lengths $I$ and $J$, respectively. 
Elements $r_i$ and $s_j$ at sequence indices $i$ and $j$ can be univariate or multivariate with dimensions $r_i=(\alpha_1, \dots, \alpha_u, \dots, \alpha_U)^{\top}$ and $s_j=(\beta_1, \dots, \beta_v, \dots, \beta_V)^{\top}$, respectively. 
Given the two sequences, $\mathbf{p}$ and $\mathbf{s}$, DTW can be used to determine the global distance between them. 
This global distance is robust to issues such as temporal distortions and has had many successes as a distance measure~\cite{ratanamahatana2004everything}. 

The key feature of DTW is that it non-linearly matches time series elements in the time dimension in order to match features and remove time distortions. 
It does this by warping the sequences so that there is an optimized alignment between elements that minimizes the global cost under constraints.  
Specifically, DTW finds the minimal path on an element-wise cost matrix $C$ using dynamic programming. 
This minimal path is referred to as the \textit{warping path} and the warping path becomes a mapping for the time steps of one series to the time steps of another. 
To solve for the minimal path, a minimal cumulative sum matrix is calculated using the recurrent function:
\begin{multline}
\label{eq:slope-symmetric}
{D}(i, j) =
\\ {C}(r_i, s_j) + \min_{(i', j') \in \{(i, j-1), (i-1, j), (i-1, j-1)\}}{D}(i', j'),
\end{multline}
where ${D}(i,j)$ is the cumulative sum of the $i$-th and $j$-th elements and ${C}(r_i, s_j)$ is the local distance between $r_u$ and $s_t$. 
In this paper, we use the Euclidean distance, or ${C}(r_i, s_j) = ||{r}_{i}-{s}_{j}||$, as the cost function. 
In the typical use of DTW, the global distance is defined as the value at ${D}(I,J)$. 
It should be noted that the slope constraint defined in Eq.~\eqref{eq:slope-symmetric} is a symmetric slope constraint~\cite{sakoe1978dynamic} and other slope constraints, such as asymmetric and weighted constraints, have been proposed~\cite{itakura1975minimum}. 
However, the proposed method can work with any variation. 
The symmetric slope constraint was selected due to being the most commonly used, but the proposed method can work with any variation of DTW. 


\subsection{Guided Warping for Data Augmentation}
\label{sec:dtwwarp}

The purpose of data augmentation for time series is to generate patterns that extend the data in order to improve generalization. 
Namely, given training set $\mathbf{S}=\left\{\mathbf{s}_{1},\dots,\mathbf{s}_{n},\dots,\mathbf{s}_{N}\right\}$ with individual time series $\mathbf{s}_{n}$, our goal is to create augmented set $\mathbf{S}'$ such that the accuracy of a model trained on $\mathbf{S}\cup\mathbf{S}'$ is greater than $\mathbf{S}$ alone.
However, the patterns of $\mathbf{S}'$ need to be similar to the original feature distribution of $\mathbf{S}$ as to not introduce too much noise or create illogical patterns.

To generate time series, we propose the use of guided warping, or using guidance to instruct warping in the time domain based on other reference patterns. 
Typically, time warping for data augmentation is done using random warping~\cite{Um_2017} or random windows~\cite{le2016data}. 
However, instead of randomly time warping, we propose creating augmentation set $\mathbf{S}'$ using patterns from training set $\mathbf{S}$ that are time warped using the guidance of other intra-class patterns from $\mathbf{S}$. 
Doing so is a form of pattern mixing where we preserve the features of student time series $\mathbf{s}$ and set it to the pace of teacher time series $\mathbf{r}$. 
The advantage of warping using a reference over randomly warping is that both the local features and the time steps they occur at exist in the original dataset. 
Randomly warping only hopes that the generated patterns are realistic.

To align the elements of $\mathbf{s}$ and $\mathbf{r}$ so guided warping can take place, we use DTW. 
As explained previously, traditionally, DTW is used as a distance measure for finding the global distance between the two. 
However, similar to other works~\cite{Petitjean_2011,Wu_2019,Iwana_2019}, we can exploit the warping path to align the elements of the sequences. 
By aligning the elements in this way, sections of $\mathbf{s}$ are warped in the time dimension to fit $\mathbf{r}$. 

Specifically, a minimal cumulative sum matrix for DTW is calculated using Eq.~\eqref{eq:slope-symmetric} and the minimum warping path is found by tracing $D(I,J)$ back to the origin $D(0,0)$ through matched elements $(i',j')$. 
Next, time series $\mathbf{s}'$ is created by warping $\mathbf{s}$ by the time steps $1,\dots,j',\dots,J'$. 
The result is a sequence $\mathbf{s}'$ that has the feature values of $\mathbf{s}$ but the time steps of $\mathbf{r}$ under the warping path constraints provided by DTW. 
Finally, the process is repeated by selecting any two random patterns in $\mathbf{S}$ that are the same class. 
Using this method, it is possible to synthesize $\sum_{y} N_{y}^2$ number of time series where $N_{y}$ is the number of patterns in each class $y$. 

\section{Improved Time Series Generation with Shape Descriptors}
\label{sec:shapedtw}

\begin{figure}[!t]
\centering
\subfloat[DTW]{\includegraphics[width=0.49\columnwidth,trim=1.2cm 1.3cm 0.4cm 1.5cm,clip]{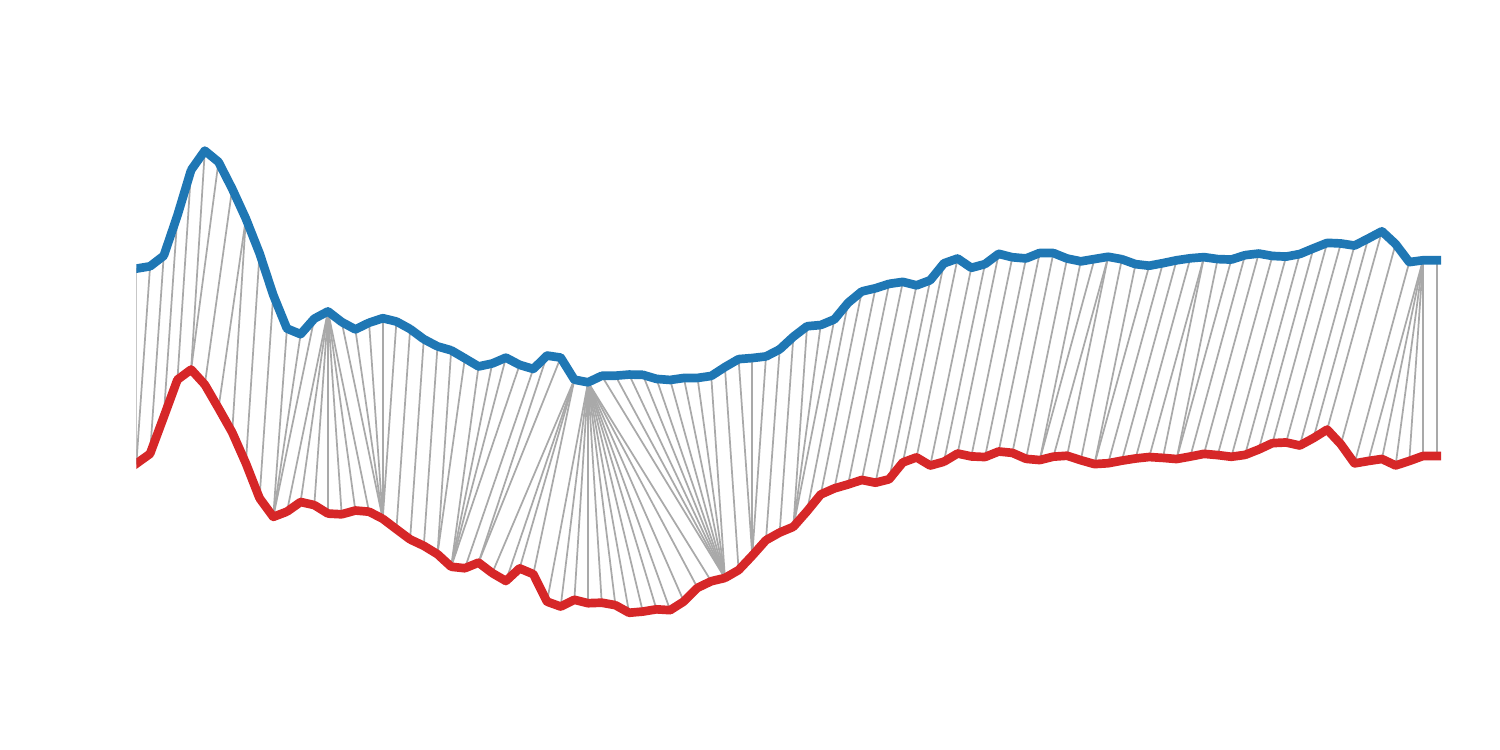}}~
\subfloat[shapeDTW]{\includegraphics[width=0.49\columnwidth,trim=1.2cm 1.3cm 0.4cm 1.5cm,clip]{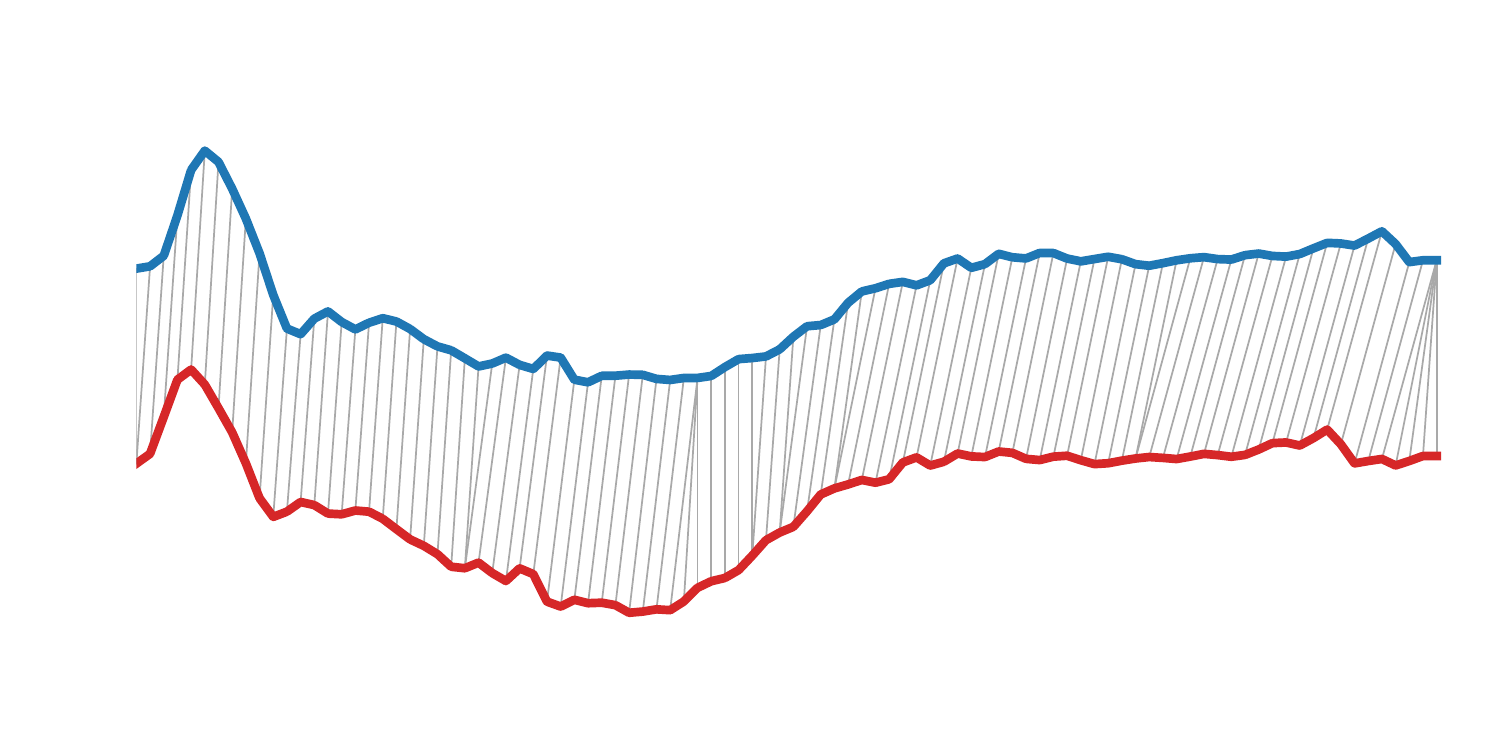}}
\caption{Comparison between alignment with (a) DTW and (b) shapeDTW. The red and blue lines are time series and the gray connections are the element alignments.
\vspace{-3mm}
}
\label{fig:dtwproblem}
\end{figure}

DTW finds the optimal alignment between individual elements by minimizing the global cost between elements on the warping path. 
While this is effective for distance measures, this might not be optimal for pattern generation. 
Using a strict minimal optimization can produce jagged and abrupt changes in the warping path, as shown in Fig.~\ref{fig:dtwproblem}. 
To overcome this problem, we propose using a method of alignment based on high-level features, namely shapeDTW~\cite{Zhao_2018}. 

Instead of using element-wise matching as in DTW, shapeDTW dynamically matches shape descriptors within the sequence. 
Given time series $\mathbf{s}$, a shape descriptor is a multivariate vector $d_{s_j}=(s_{j-\lceil\frac{1}{2}W\rceil},\dots,s_{j},\dots,s_{j+\lfloor\frac{1}{2}W\rfloor})^{\top}$ created from a subsequence of $\mathbf{s}$ with a length of $W$, centered on element $j$. 
Using a stride of 1, a new sequence of shape descriptors $\mathbf{d_{s}}=d_{s_1},\dots,d_{s_j},\dots,d_{s_J}$ is created of equal length to the original $\mathbf{s}$. 
Also, padding is added for $j-\lceil\frac{1}{2}W\rceil<1$ and $j+\lfloor\frac{1}{2}W\rfloor>J$ with duplicates of $s_1$ and $s_J$, respectively. 
These shape descriptors represent higher-level features (i.e. segments of a time series) than the individual elements themselves.

ShapeDTW proceeds with a standard DTW calculation, however, with shape descriptor sequences $\mathbf{d_{s}}$ and $\mathbf{d_{p}}$ instead of the raw elements. 
In other words, time series $\mathbf{s}$ and $\mathbf{p}$ are temporally aligned using their shape descriptors. 
Essentially, instead of the element-wise cost function ${C}(p_i, s_j)=||p_i - s_j||$ in Eq.~\eqref{eq:slope-symmetric} of DTW, shapeDTW uses a cost function ${C}(p_i, s_j)=||d_{p_i} - d_{s_j}||$ between the shape descriptors. 
The result is a non-linear alignment much like DTW, but using the similarity between neighboring points to the elements. 

\begin{figure}[!t]
\centering
\subfloat[Warping with DTW]{\includegraphics[width=0.45\columnwidth,trim=0.4cm 0.4cm 0.3cm 0.3cm,clip]{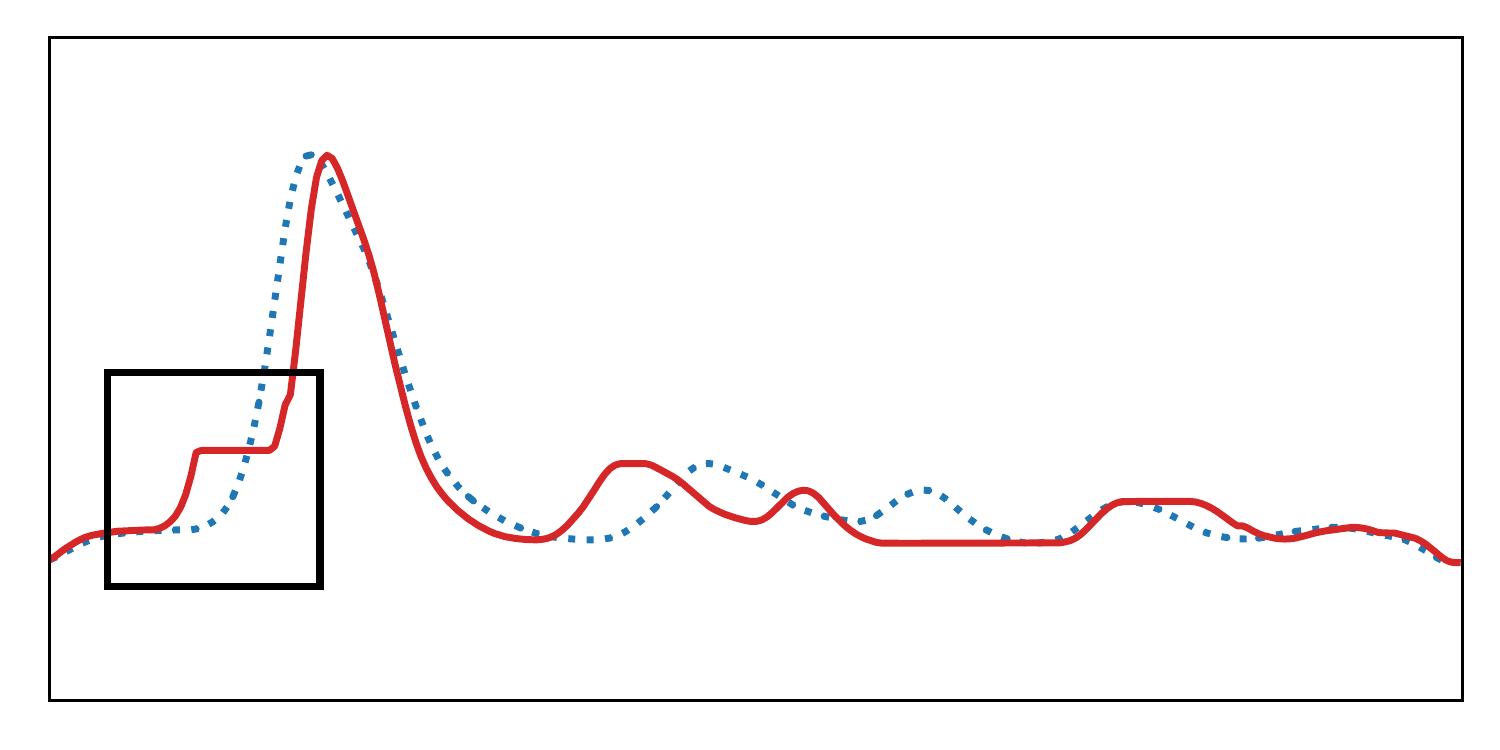}}~
\subfloat[Warping with shapeDTW]{\includegraphics[width=0.45\columnwidth,trim=0.4cm 0.4cm 0.3cm 0.3cm,clip]{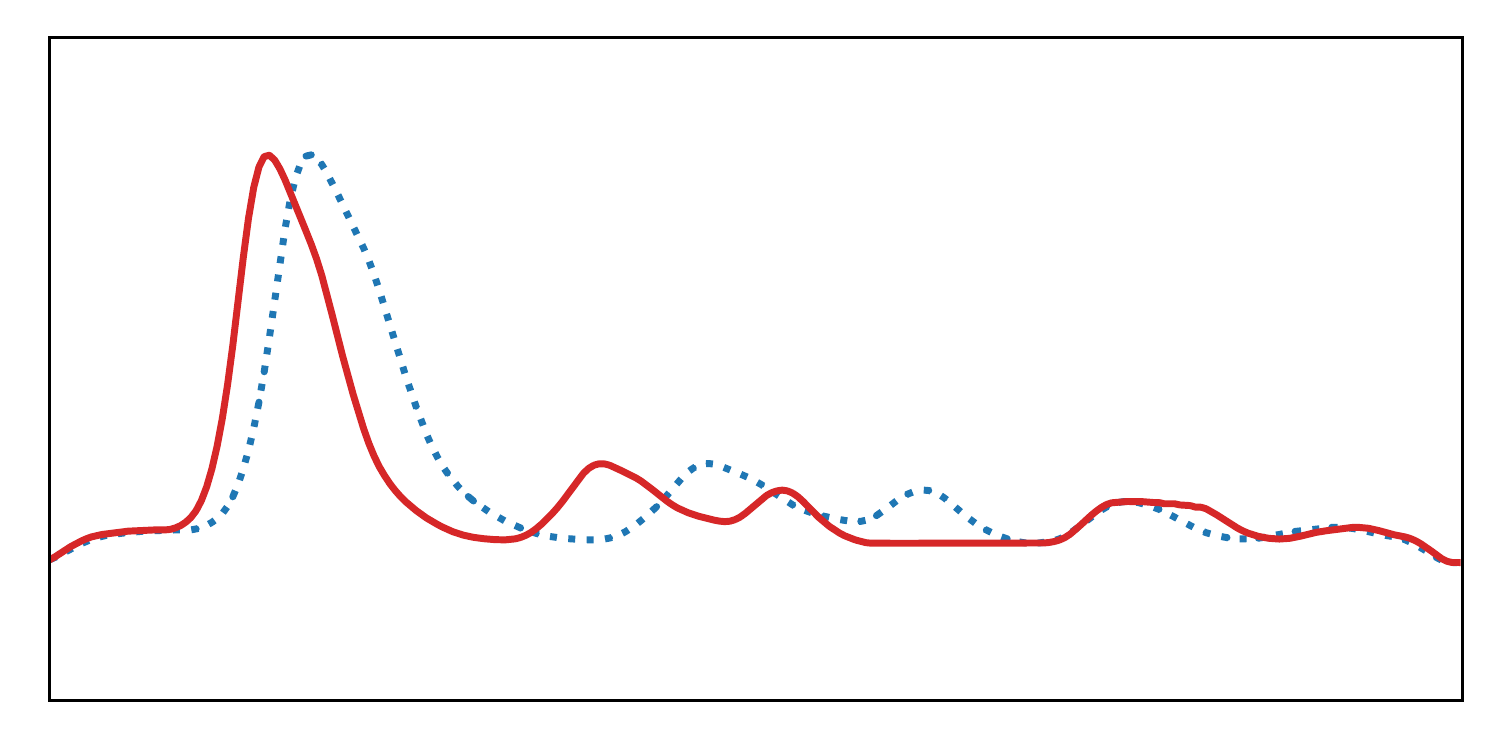}}
\caption{The difference between (a) warping using DTW and (b) warping using shapeDTW. The boxed region in (a) highlights an unnatural feature created from the dynamic alignment.
\vspace{-3mm}
}
\label{fig:example_shape}
\end{figure}

Similar to Section~\ref{sec:dtwwarp}, we can exploit the alignment byproduct of shapeDTW to generate new time series. 
Using shapeDTW, a new pattern is created by warping the features of $\mathbf{s}$ to the time steps of $\mathbf{p}$. 
Compared to guided warping with standard DTW, we are able to preserve the features of the original pattern and create more natural time series. 
For example, in Fig.~\ref{fig:example_shape}, the boxed region contains an abnormal feature that caused by overzealous minimization of the warping path by DTW. 

\section{Discriminative Teacher Selection}
\label{sec:disc}
While the reference prototype $\mathbf{r}$ can be chosen at random (within the same class), we posit that using a directed prototype is better than a random one. 
In this case, we propose to select the \textit{most discriminative} prototype from a bootstrap set of samples to be the reference, for DTW. 
The bootstrap set is used to represent the distribution of the training samples with $B\ll N$ patterns selected at random with replacement, where $B$ is the number of bootstrap samples and $N$ is the total training set size.




\subsection{Formulation}

To determine the most discriminative teacher from the bootstrap set, we use a nearest centroid classifier based on DTW (or shapeDTW) distance~\cite{iwana2017efficient}. 
Specifically, for each time series $\mathbf{b}_m$ in $\mathbf{B}\subset\mathbf{S}$, with subset size $M$:
\begin{multline}
\label{eq:weaklearner}
h({\mathbf{b}_m}) = \frac{1}{\sum_{m'}{[l_{m'} \neq l_m]}} \sum_{m' }{\mathcal{D}(\mathbf{b}_{m'}, \mathbf{b}_{m})|[l_{m'} \neq l_m]} 
\\ -\frac{1}{\sum_{m'}{[l_{m'} = l_m]}} \sum_{m'}{\mathcal{D}(\mathbf{b}_{m'}, \mathbf{b}_{m})|[l_{m'} = l_m]},
\end{multline}
where $l_{m'}$ is the label for each $\mathbf{b}_m$ and the hypothesis is $\sign(h({\mathbf{b}_{m}}))$. 
Due to the high possibility of ties, instead of using the prediction $\sign(h({\mathbf{b}_{m}}))$, we use the reference with the maximal distance between the positive and negative centroids, in other words:
\begin{equation}
    \mathbf{b}_{\mathrm{disc}} = \argmax_{\left\{m=1,\dots, M\right\}} h({\mathbf{b}_m}), 
\end{equation}
where $\mathbf{b}_{\mathrm{disc}}$ is the reference that we define as the most discriminative in $\mathbf{B}$.
Also, it should be noted that to ensure successful selection, we sample from $\mathbf{S}$ as close to evenly for $\mathbf{B}$ from same class patterns $l_{m'} = l_m$ and different class patterns $l_{m'} \neq l_m$ as possible.
Using the selected discriminative teacher $\mathbf{b}_{\mathrm{disc}}$, $\mathbf{s}$ is warped using DTW (or shapeDTW) as described previously. 
We distinguish the two proposed methods by referring to the random selection as Random Guided Warping~(RGW) and the discriminative teacher selection as Discriminative Guided Warping~(DGW) with the DTW variants (-D) referred to as RGW-D and DGW-D and the shapeDTW variants (-sD) as RGW-sD and DGW-sD, respectively.

\begin{figure}[!t]
\centering
\subfloat[RGW-D (Proposed)]{\fbox{\includegraphics[width=0.29\columnwidth,trim=1.3cm 1.2cm 0.7cm 0.7cm,clip]{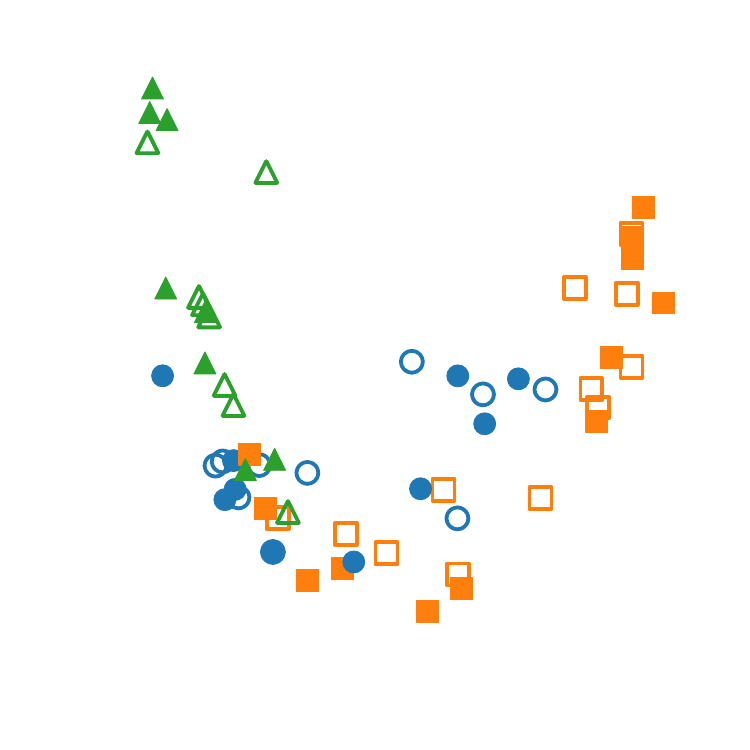}}}~
\subfloat[RGW-sD (Proposed)]{\fbox{\includegraphics[width=0.29\columnwidth,trim=1.3cm 1.2cm 0.7cm 0.7cm,clip]{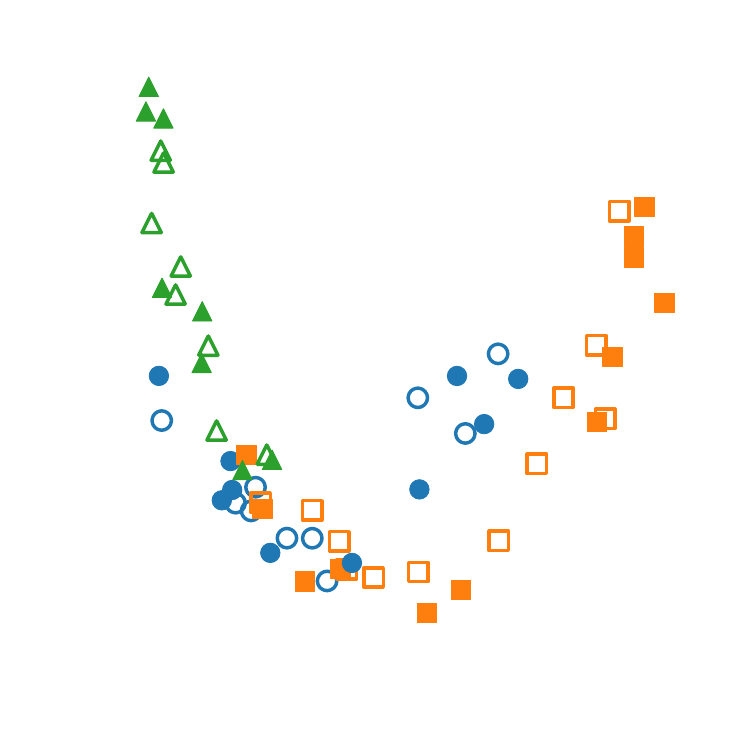}}}~
\subfloat[DGW-D (Proposed)]{\fbox{\includegraphics[width=0.29\columnwidth,trim=1.3cm 1.2cm 0.7cm 0.7cm,clip]{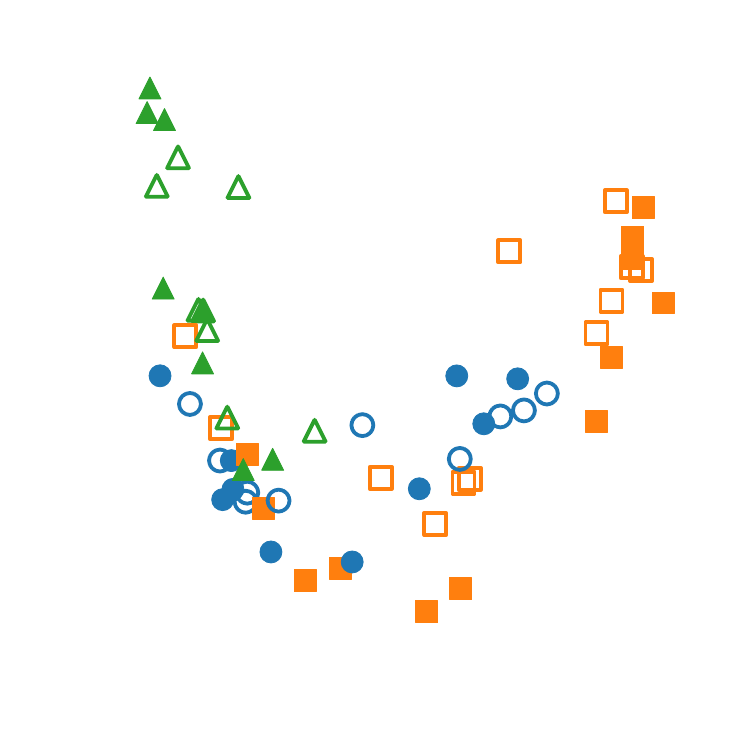}}}

\subfloat[DGW-sD (Proposed)]{\fbox{\includegraphics[width=0.29\columnwidth,trim=1.3cm 1.2cm 0.7cm 0.7cm,clip]{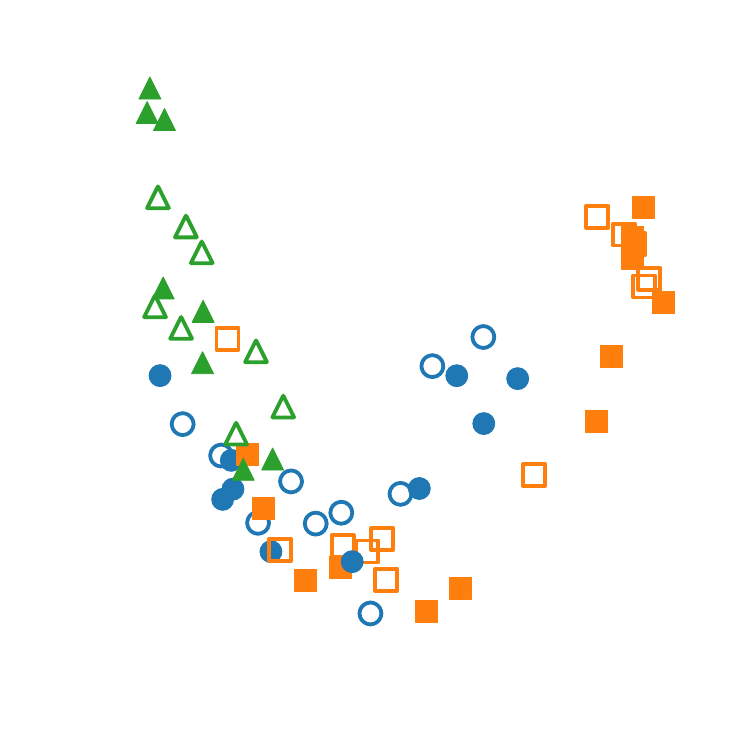}}}~
\subfloat[SPAWNER~\cite{Kamycki_2019}]{\fbox{\includegraphics[width=0.29\columnwidth,trim=1.3cm 1.2cm 0.7cm 0.7cm,clip]{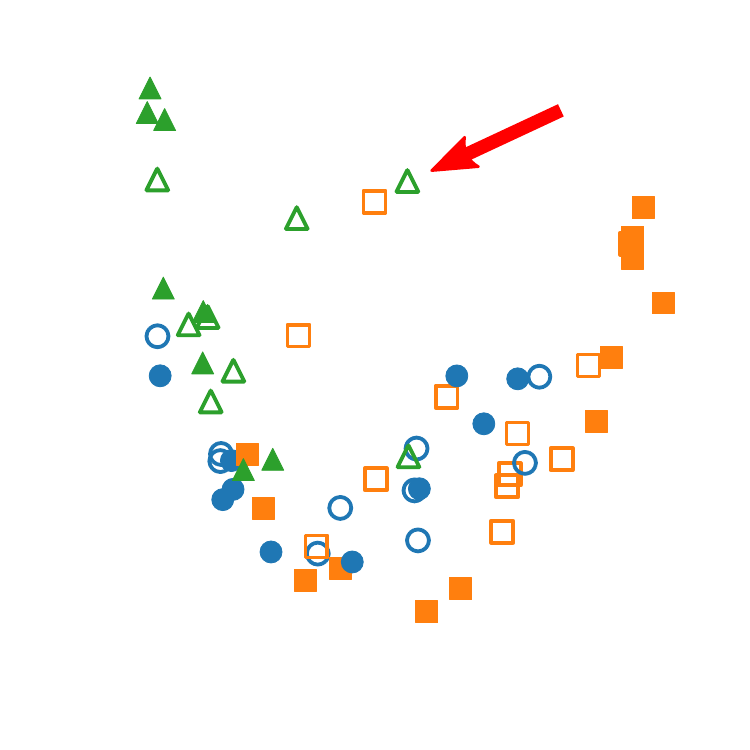}}}~
\subfloat[wDBA ASD~\cite{Forestier_2017}]{\fbox{\includegraphics[width=0.29\columnwidth,trim=1.3cm 1.2cm 0.7cm 0.7cm,clip]{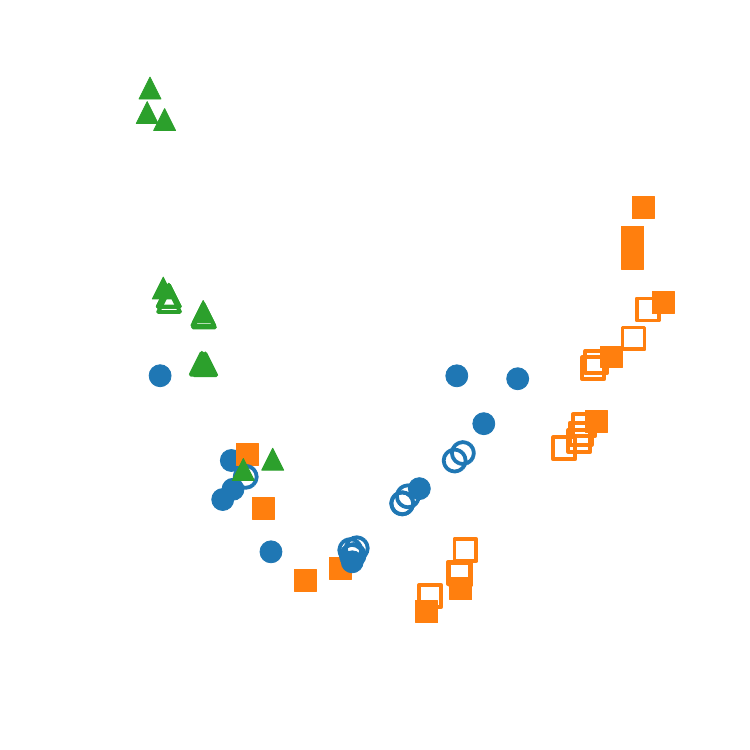}}}
\caption{Visualization of the CBF dataset using PCA using the proposed augmentation and the other pattern mixing methods. The solid shapes are the original time series and the hollow shapes are the generated time series. The arrow points to examples generated patterns that do not fit the expected distribution of features in the dataset.
\vspace{-3mm}
}
\label{fig:pca}
\end{figure}

\subsection{Visualization}

To demonstrate the effects of the proposed selection process, the data can be visualized using Principal Component Analyis~(PCA)~\cite{Pearson_1901}. 
For Fig.~\ref{fig:pca}, CBF is selected as an example time series dataset due to having a very small amount of samples (only 30). 
In the figure, each color represents one class with the solid shapes being original samples and the hollow shapes being generated examples. 

The figure highlights the differences between the proposed methods and the two other DTW based pattern mixing data augmentation methods. 
Since RGW (Fig.~\ref{fig:pca}~(a)) and SPAWNER~\cite{Kamycki_2019} (Fig.~\ref{fig:pca}~(e)) use randomly selected references, it is possible to generate patterns that seem to not fit with the data, such as the patterns in the center of the ``U'' shape. 
Using a discriminative teacher (Fig.~\ref{fig:pca}~(c)) can help direct the warping toward useful patterns. 
Furthermore, incorporating shapeDTW helps both methods in maintaining more consistency in the data distribution.
As for wDBA ASD~\cite{Forestier_2017} (Fig.~\ref{fig:pca}~(f)), the generated patterns are too similar to existing patterns to be useful for data augmentation. 
Using shapeDTW with DGW, in Fig.~\ref{fig:pca}~(d), provides the best balance of maintaining the data distribution while producing new patterns.

\begin{figure}[!t]
\centering
OliveOil\vspace{1mm}
\subfloat[DGW-sD (Proposed)]{\includegraphics[width=0.32\columnwidth,trim=0.4cm 0.4cm 0.3cm 0.3cm,clip]{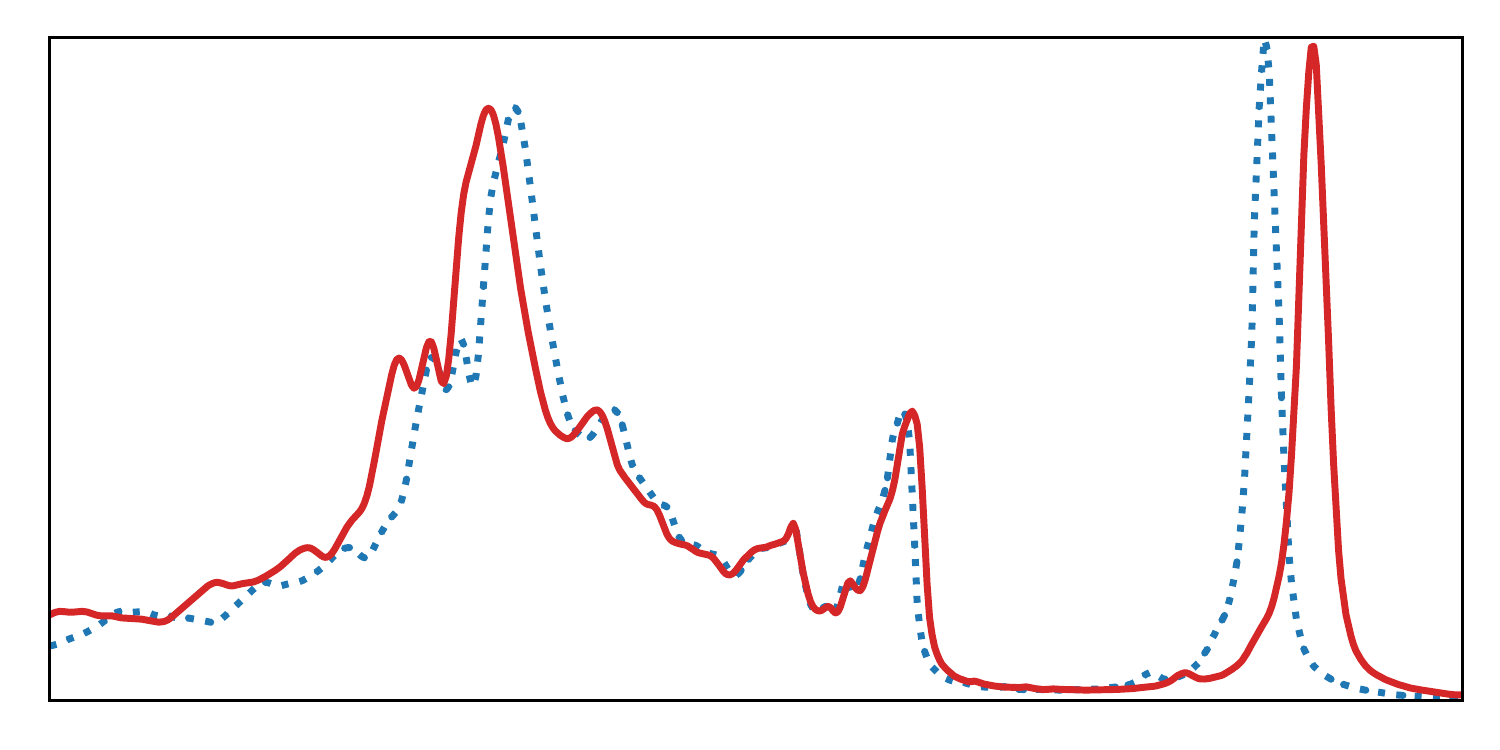}}~
\subfloat[SPAWNER~\cite{Kamycki_2019}]{\includegraphics[width=0.32\columnwidth,trim=0.4cm 0.4cm 0.3cm 0.3cm,clip]{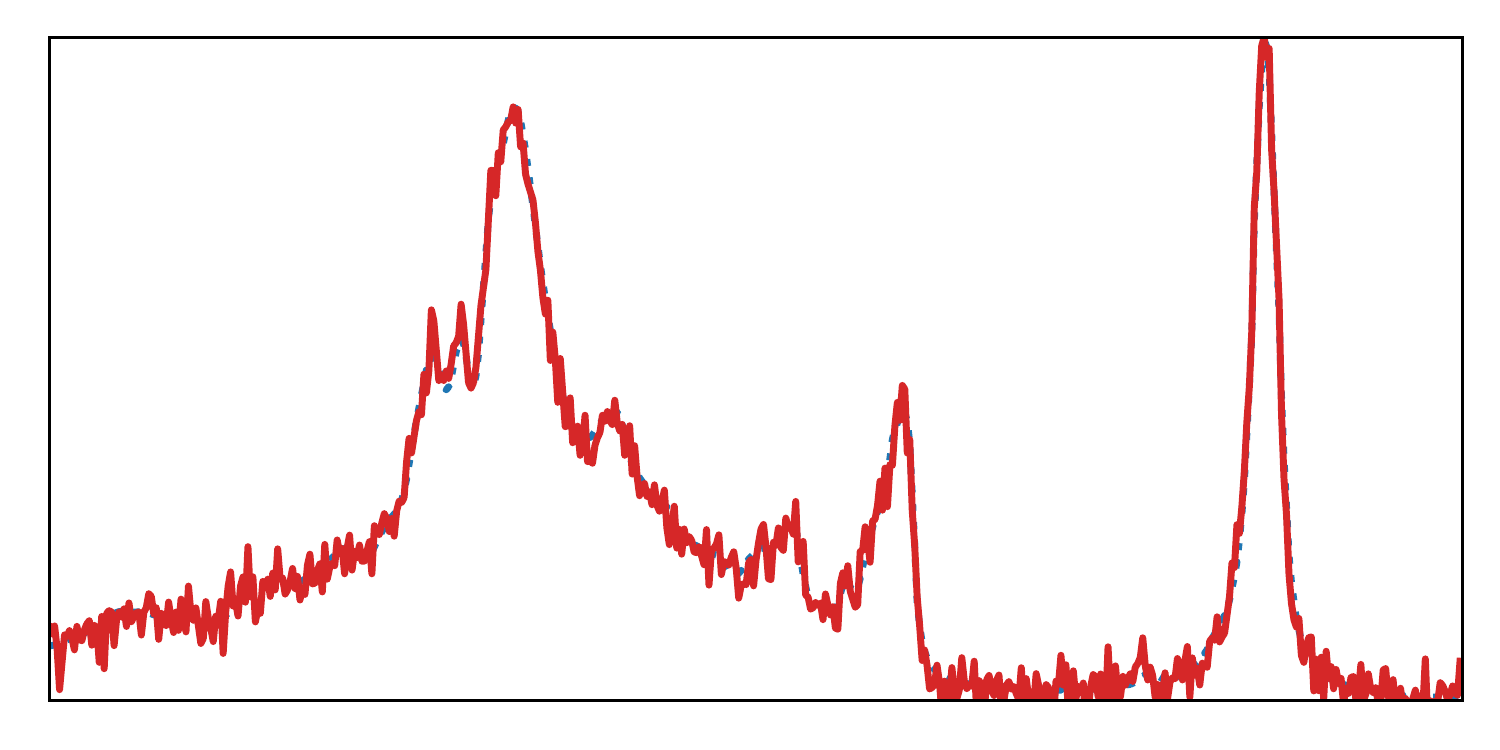}}~
\subfloat[wDBA ASD~\cite{Forestier_2017}]{\includegraphics[width=0.32\columnwidth,trim=0.4cm 0.4cm 0.3cm 0.3cm,clip]{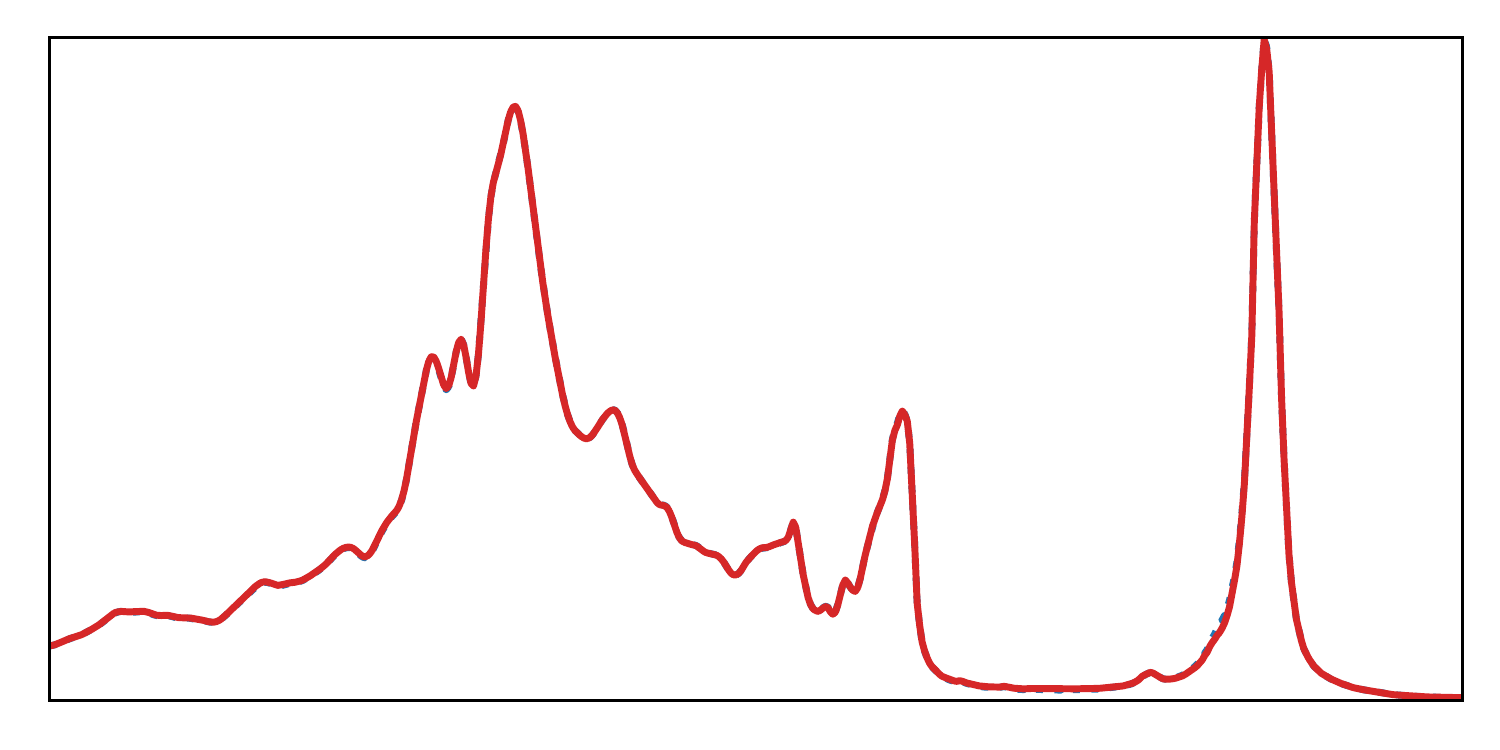}}

\vspace{2mm}
Symbols
\subfloat[DGW-sD (Proposed)]{\includegraphics[width=0.32\columnwidth,trim=0.4cm 0.4cm 0.3cm 0.3cm,clip]{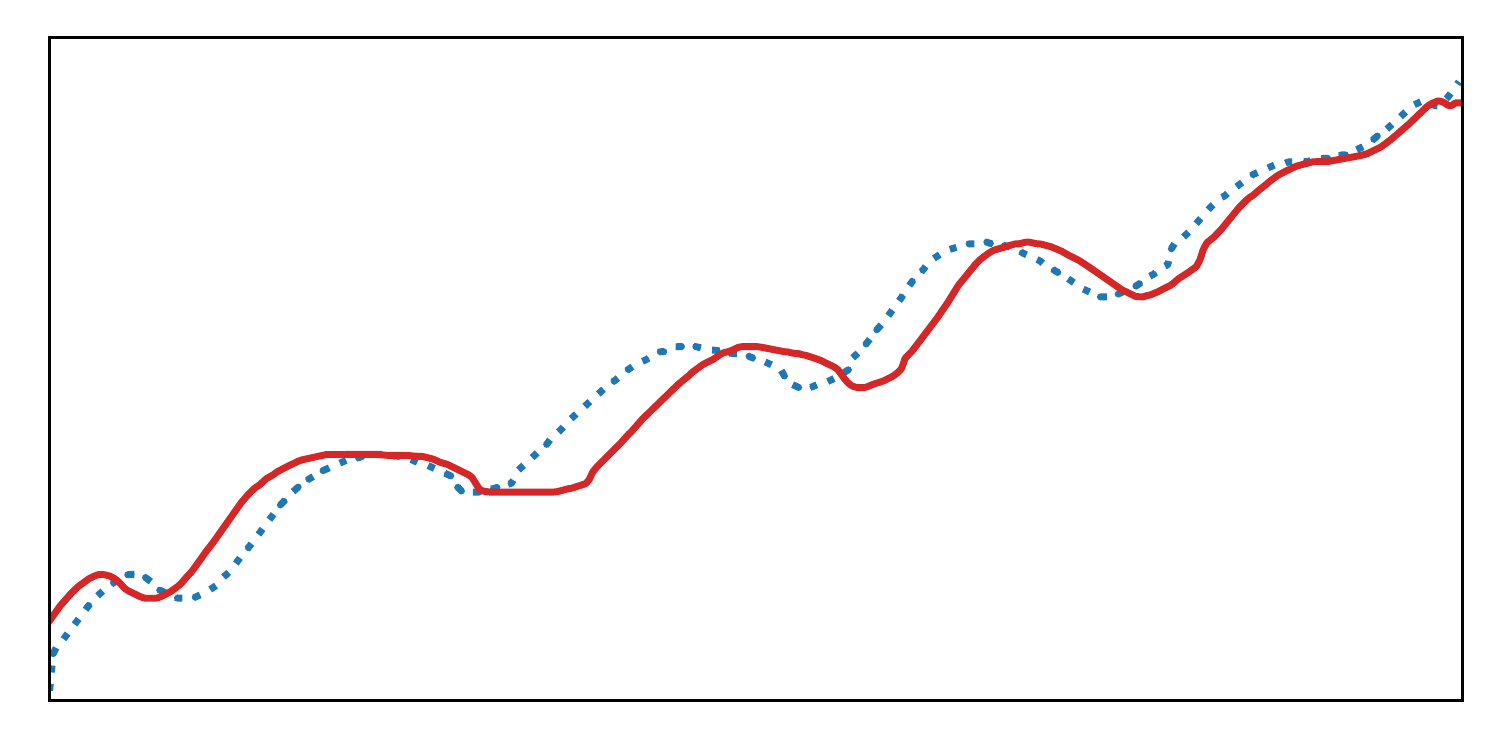}}~
\subfloat[SPAWNER~\cite{Kamycki_2019}]{\includegraphics[width=0.32\columnwidth,trim=0.4cm 0.4cm 0.3cm 0.3cm,clip]{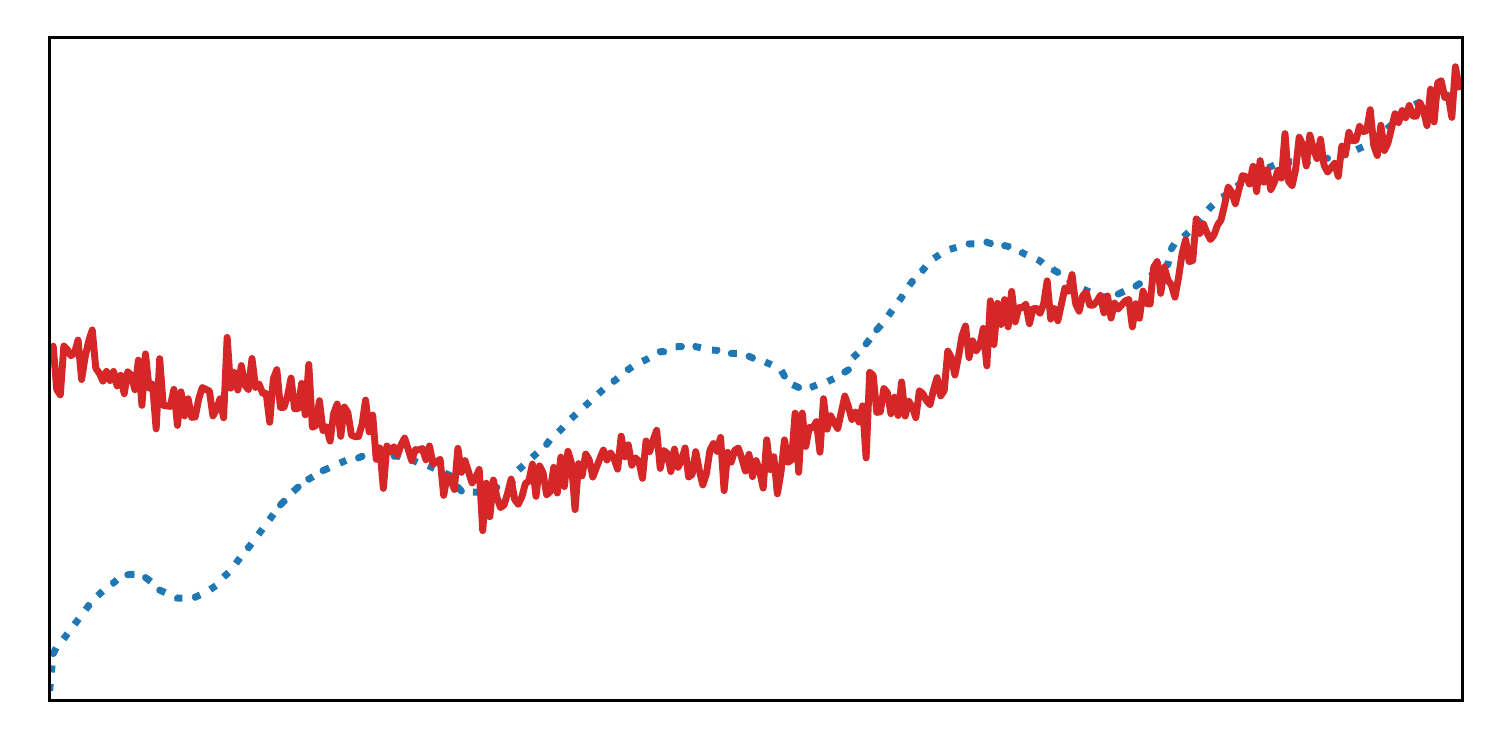}}~
\subfloat[wDBA ASD~\cite{Forestier_2017}]{\includegraphics[width=0.32\columnwidth,trim=0.4cm 0.4cm 0.3cm 0.3cm,clip]{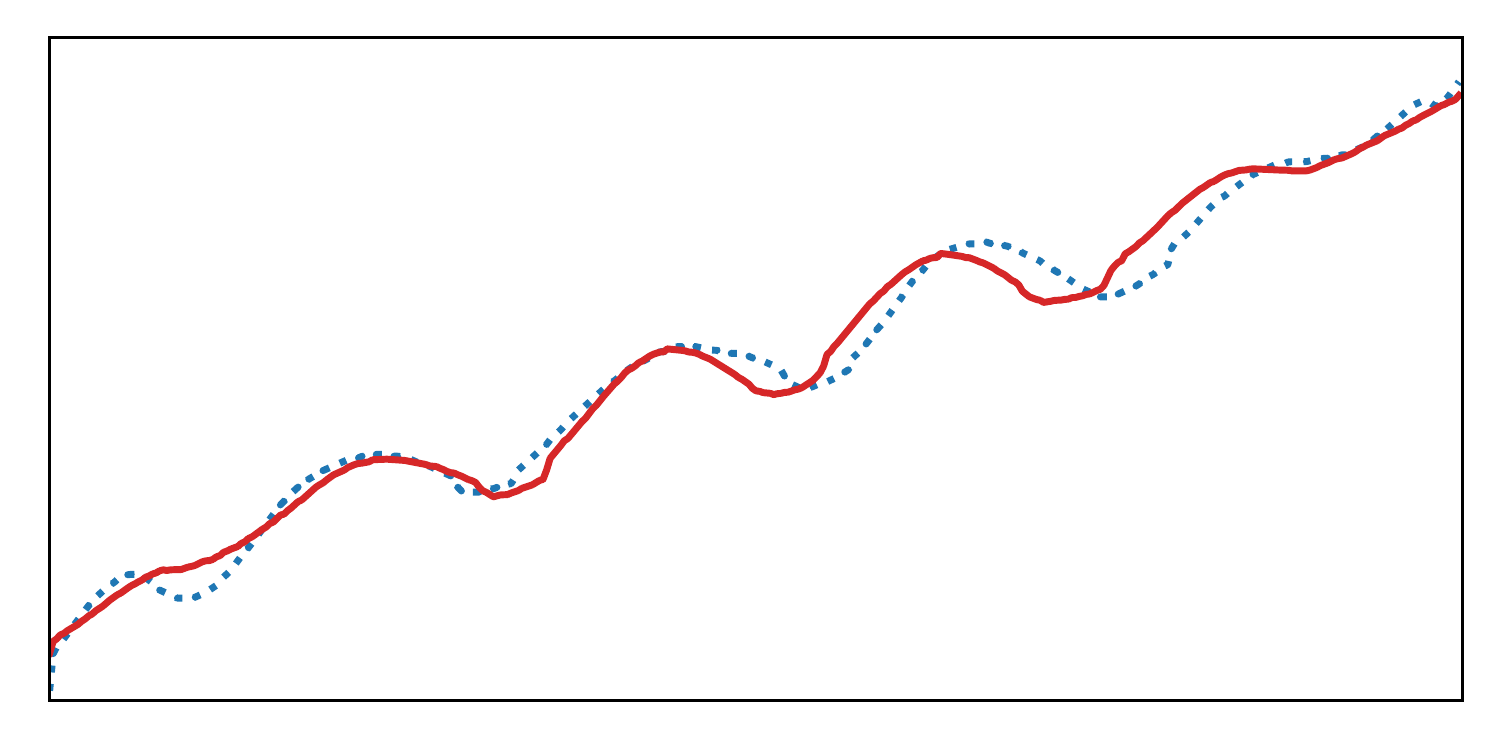}}

\caption{Examples of generated patterns. The blue dotted line is the original time series and the red solid lines are the generated time series.
\vspace{-3mm}
}
\label{fig:gen}
\end{figure}

Specific examples of these observations can be seen in Fig.~\ref{fig:gen}. 
In Fig.~\ref{fig:gen} (c) and (f), The time series generated by wDBA are too similar to the original time series. 
Whereas, SPAWNER created patterns that are not typical to the dataset. 
Especially in the case of Fig.~\ref{fig:gen} (e), where two patterns of the same class that were drastically different from each other were selected. 
DGW-sD was able to generate time series that fit the distribution of the dataset but are different enough to be useful for augmentation. 

\section{Experimental Results}
\label{sec:results}

\subsection{Datasets}

We want to assess the effects that the proposed method and the other data augmentation techniques have on a wide variety of datasets. 
Therefore, we used all 85 datasets of the 2015 UCR Time Series Archive~\cite{UCRArchive}. 
The datasets are a collection of 6 electric device, 6 ECG, 29 image outline, 14 motion capture, 18 sensor reading, 5 simulated, and 7 spectrograph time series. 
They have fixed training and test sets with between 16 and 8,926 training samples and between 20 and 8,236 test samples. 
Furthermore, the time series lengths range between 24 and 2,709 time steps. 
For pre-processing, each dataset is normalized so that the largest and smallest values in the training dataset is 1 and -1, respectively. 

\begin{table*}[!t]
\setlength{\tabcolsep}{5pt}
\renewcommand{\arraystretch}{1.3}
\caption{Test Accuracy (\%) Grouped by Time Series Type}
\label{tab:vggresults}
\centering
\begin{tabular}{l|cccccccccc|cccc}
\hline
Type & &\multicolumn{8}{c}{Comparative Augmentation with VGG} && \multicolumn{4}{c}{Proposed Augmentation with VGG} \\
(count) & None & Jit & Rot & Scal & MagW & TimW & Slic & WinW & SPAWNER & wDBA & RGW-D & RGW-sD& DGW-D & DGW-sD \\
\hline
Device (6)      & 54.9 & 54.9 & 56.4 & 57.5 & 56.4 & 58.7 & \textbf{62.1} & 57.6 & 58.5 & 56.5 & 60.4 & 59.8 & 60.0 & 59.5  \\
ECG (6)         & 93.7 & 93.7 & 93.6 & 93.8 & \textbf{93.9} & 90.2 & 93.1 & 93.0 & 92.9 & 88.4 & 92.8 & 91.9 & 92.2 & 92.2  \\
Image (29)      & 75.4 & 78.3 & 78.1 & 74.3 & 78.6 & 79.8 & 80.2 & \textbf{81.5} & 78.8 & 78.2 & 80.1 & 80.3 & 81.0 & 81.1  \\
Motion (14)     & 70.7 & 69.9 & 66.7 & 68.7 & 70.3 & 73.1 & 70.1 & 72.7 & 71.3 & 71.8 & 72.3 & 73.2 & \textbf{73.9} & 73.7  \\
Sensor (18)     & 80.3 & 80.4 & 79.7 & 79.3 & 80.4 & 79.5 & 79.9 & 81.2 & 80.2 & 79.7 & 80.3 & 80.8 & \textbf{81.7} & 80.8  \\
Sim. (5)   & 89.1 & 89.1 & 73.7 & 86.8 & 88.3 & 91.0 & 94.8 & 96.6 & 96.5 & 87.9 & 89.4 & 89.5 & 91.9 & \textbf{98.6} \\
Spectro (7)     & 76.9 & 77.3 & 81.5 & 79.0 & \textbf{86.0} & 74.4 & 82.5 & 76.5 & 83.5 & 80.7 & 85.9 & 81.4 & 83.5  & 81.7 \\
\hline
Total (85)           & 76.44 & 77.32 & 74.84 & 77.06 & 78.30 & 78.10 & 79.15 & {79.58} & 78.84 & 77.42 & 79.39 & 79.27 & 80.12 & \textbf{80.17}\\
\hline
Type & &\multicolumn{8}{c}{Comparative Augmentation with LSTM} && \multicolumn{4}{c}{Proposed Augmentation with LSTM} \\
(count) & None & Jit & Rot & Scal & MagW & TimW & Slic & WinW & SPAWNER & wDBA & RGW-D & RGW-sD& DGW-D & DGW-sD \\
\hline
Device (6)      & 40.8 & 43.0 & 41.1 & 41.7 & 43.4 & 43.5 & \textbf{44.6} & 42.5 & 42.5 & 42.5 & 44.2 & 44.3 & \textbf{44.6} & 42.3\\
ECG (6)         & 57.6 & 70.3 & 61.1 & \textbf{78.7} & 63.4 & 49.2 & 61.0 & 55.1 & 65.0 & 59.5 & 61.1 & 55.6 & 59.6 & 59.1\\
Image (29)      & 62.2 & 61.5 & 59.4 & 57.8 & 63.5 & 54.7 & 55.0 & 62.4 & \textbf{63.6} & 58.8 & 63.2 & 62.0 & 57.5 & 60.1\\
Motion (14)     & 45.0 & 45.5 & 37.7 & 42.6 & 43.8 & 41.4 & 40.1 & \textbf{48.4} & 43.1 & 44.6 & 38.4 & 43.5 & 43.7 & 43.9\\
Sensor (18)     & 59.2 & 62.2 & 58.1 & 61.5 & 61.4 & 57.6 & 61.5 & 60.3 & 62.0 & 61.4 & \textbf{63.2} & 61.4 & 62.9 & 63.1\\
Sim. (5)        & 72.6 & 70.5 & 68.6 & \textbf{76.6} & 72.8 & 72.7 & 63.7 & 68.8 & 76.1 & 74.2 & 72.3 & 62.2 & 63.6 & 73.8\\
Spectro (7)     & 58.9 & 55.4 & 57.9 & \textbf{61.6} & 54.1 & 52.5 & 55.6 & 52.8 & 60.5 & 49.0 & 54.1 & 53.4 & 58.3 & 53.3\\
\hline
Total (85)      & 57.24 & 58.35 & 54.78 & 57.98 & 58.04 & 52.80 & 54.08 & 57.49 & \textbf{58.98} & 56.01 & 57.42 & 56.43 & 56.01 & 56.99 \\
\hline
\end{tabular}
\vspace{-3mm}
\end{table*}

\subsection{Augmentation Methods}

To evaluate the proposed method, we used nine general time series data augmentation techniques found from literature. 
The parameters of each of the comparison augmentation methods were set to be the same as used by the respective works.
The following comparison methods were used for evaluations:
\begin{itemize}
    \item \textbf{None}: Uses no augmentation for a baseline.
    \item \textbf{Jittering (Jit)}: Random noise from a Gaussian distribution with a mean $\mu=0$ and a standard deviation $\sigma=0.03$, as suggested in~\cite{Um_2017}, is added to the original time series.
    \item \textbf{Rotation (Rot)}: For rotation, since the patterns in the UCR Time Series Archive are univariate, patterns are randomly flipped.
    \item \textbf{Scaling (Scal)}: In scaling, the magnitude of all elements in the time series is increased or decreased by a scalar. As in~\cite{Um_2017}, the scalar is determined by a Gaussian distribution with $\mu=1$ and $\sigma=0.1$.
    \item \textbf{Magnitude Warping (MagW)}: The magnitude of each time series is multiplied by a curve created by cubic spline with four knots at random magnitudes with $\mu=1$ and $\sigma=0.2$~\cite{Um_2017}.
    \item \textbf{Time Warping (TimW)}: Time warping based on a random smooth warping curve generated by cubic spline with four knots at random magnitudes ($\mu=1$, $\sigma=0.2$). We followed the same procedure as Um et al.~\cite{Um_2017}
    \item \textbf{Slicing (Slic)}: For this augmentation method, we use window slicing~\cite{le2016data}. In window slicing, a window of 90\% of the original time series is chosen at random. In our implementation, we interpolate this back to the original size to fit with the classifier.
    \item \textbf{Window Warping (WinW)}: Window warping~\cite{le2016data} selects a random window of 10\% of the original data and either speeds it up by $2$ or slows it down by $0.5$. 
    \item \textbf{Suboptimal Warped Time Series Generator (SPAWNER)}: SPAWNER~\cite{Kamycki_2019} is a pattern mixing method that creates a time series from the average of two random suboptimally aligned intra-class patterns. Furthermore, as recommended, noise is added to the average with a $\sigma=0.5$ in order to avoid instances where there is very little change.
    \item \textbf{wDBA}: We use the Average Selected with Distance~(ASD) version due to it having the best results in~\cite{Forestier_2017}. In this data augmentation method, 6 patterns weighted by their DTW distance to the medoid are averaged using DBA. 
\end{itemize}

As for the proposed method, we used the following evaluations:
\begin{itemize}
    \item \textbf{Random Guided Warping with DTW (RGW-D)}: RGW-D is the proposed method described in Section~\ref{sec:dtwwarp} which selects two intra-class patterns and warps the features of one pattern by the time steps of the second.
    \item \textbf{Random Guided Warping with shapeDTW (RGW-sD)}: This evaluation is used to show the effects of using shapeDTW to warp the high-level features. It follows the same procedure as RGW-D but with shapeDTW. The length of the shape descriptor was set to $W=\frac{1}{20}\times J$ with a minimum of $W=5$ and a maximum of $W=100$.
    \item \textbf{Discriminative Guided Warping with DTW (DGW-D)}: This is the proposed augmentation process of using a discriminative teacher as the reference for guided warping, as described in Section~\ref{sec:disc}. We use $M=6$ for the bootstrap batch size.
    \item \textbf{Discriminative Guided Warping with shapeDTW (DGW-sD)}: DGW-sD is DGW-D but with shapeDTW as the distance measure.
\end{itemize}

For each data augmentation technique, each training set is augmented with $4\times$ the original set size. 
Also, for the methods that incorporate DTW into their algorithms, we used the symmetric slope constraint defined by Eq.~\eqref{eq:slope-symmetric} with a warping window of 10\% of the original time series length.

\subsection{Evaluation Settings}

The proposed method is evaluated on two established temporal neural network models, a VGG~\cite{simonyan2014very} and an LSTM~\cite{Hochreiter_1997}. 
A separate network is trained and tested using each dataset with each augmentation technique and one extra with no augmentation. 
They are all trained with a fixed 10,000 iterations in order to have a fair comparison between the difference between no augmentation and having augmentation. 

The VGG used in our evaluation is modified for time series by using 1D convolutions and max pooling instead of the traditional 2D convolutions and pooling for images. 
It consists of multiple blocks of convolutional layers followed by max pooling layers, two fully-connected layers of 4,096 nodes with a dropout probability of 0.5, and an output layer. 
Each layer has a Rectified Linear Unit~(ReLU) activation function except the output which uses softmax. 
The first two blocks of convolutional layers are made of two consecutive convolutional layers with 64 and 128 nodes, respectively, and the subsequent blocks have three consecutive convolutional layers of 256 nodes for the third and 512 nodes, thereafter. 
Due to the differences in the sequence lengths of the datasets, we use a varying number of blocks in order to prevent excessive pooling. 
The number of blocks $G$ used is:
\begin{equation}
    \label{eq:numpool}
    G = \textrm{round}(\log_2(J))-3,
\end{equation}
where $J$ is the number of time steps. 
This scheme keeps the output of the final max pooling to between 5 and 12 time steps~\cite{Kenji_Iwana_2020}.
For training, we use Stochastic Gradient Decent~(SGD) with an initial learning rate of 0.01, momentum of 0.9, and weight decay of $5\times10^{-4}$. 
The learning rate is reduced by 0.1 upon training accuracy plateau. 
In addition, the VGG is trained with mini-batches of 256. 
These settings are used to match~\cite{simonyan2014very}.

For the LSTM, we use some of the recommendations from Reimers et al.~\cite{reimers2017optimal}. 
Specifically, we use a two-layer LSTM with 100 cells each. 
As following~\cite{reimers2017optimal}, we use an Nadam~\cite{dozat2016incorporating} optimizer with an initial learning rate of 0.001 and mini-batch size 32.
We also use a learning rate reduction of 0.1 upon training accuracy plateau for the LSTM.

\begin{figure}[!t]
\centering
\includegraphics[width=0.62\columnwidth,trim=0.2cm 0.4cm 0.3cm 0.4cm,clip]{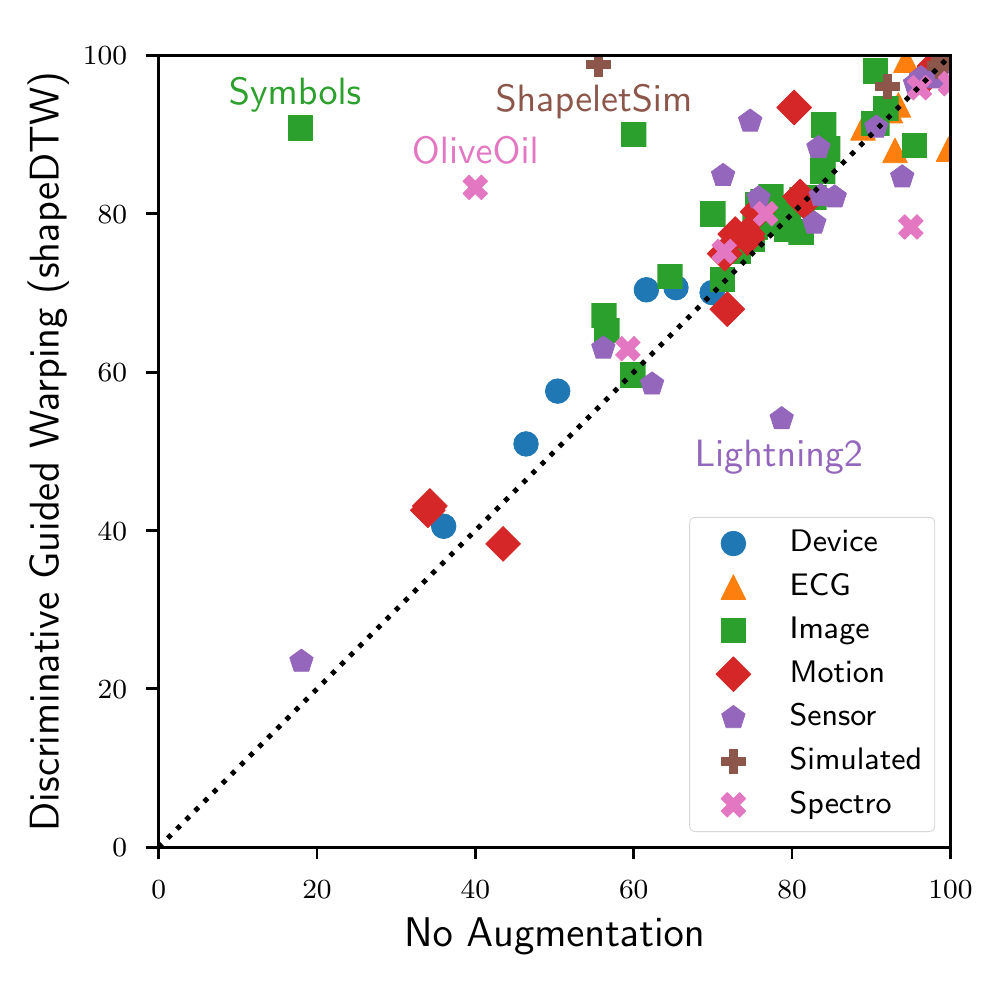}
\caption{A comparison of no augmentation and the proposed method with a VGG. Each point is a dataset and the shapes indicate dataset type.
\vspace{-3mm}
}
\label{fig:comp_noaug}
\end{figure}

\subsection{Data Augmentation Comparison Results}
The results comparing the data augmentation techniques are shown in Table~\ref{tab:vggresults}. 
For VGG, there are significant improvements using the proposed data augmentation for most of the datasets. 
Overall, there was an average of 3-4\% increase in accuracy for all of the guided warping methods.
Of the proposed methods DGW-sD had the highest overall. 
In addition, using the discriminative prototype as a reference versus a random reference improved the accuracy for both DTW and shapeDTW. 
The highest gains for DGW-sD were from Simulated, Image outline, and Device time series with increases of accuracy of 9.5\%, 6.2\%, and 5.1\%, respectively. 
However, there were dataset categories that did not improve. 
ECG had a slight degradation in accuracy and Sensor had insignificant improvements. 
Although, it is notable that ECG and Sensor datasets were not significantly affected by any of the data augmentation methods with VGG.

On the other hand, from Table~\ref{tab:vggresults}, there is no clear trend or advantage in using the proposed method with an LSTM. 
In general, the time domain augmentations (TimW, Slic, WinW, and the proposed methods) tended not to improve the trained LSTM models. 
This is due to LSTMs, and RNNs in general, being designed to explicitly combat time distortions. 
Thus, it is more recommended to use the proposed method with CNN based architectures.



The individual differences between the accuracies of each dataset for no augmentation and DGW-sD for VGG is shown in Fig.~\ref{fig:comp_noaug}. 
The figure shows which kind of datasets improved with DGW-sD data augmentation. 
As stated previously, image outlines, device, and simulated datasets improved the most. 
Notably, Symbols, OliveOil, and ShapeletSim had the largest increase in accuracy over no augmentation. 
For some of these datasets, we can expect increases in accuracy. 
Symbols is made of online hand-drawn symbols so maintaining the shape structure is important. 
In addition, ShapeletSim is created from shapelets embedded within time series, thus, using guided warping could help increase the generalization by moving the shapelets around in the time dimension.

\begin{figure}[!t]
\centering
\includegraphics[width=1\columnwidth,trim=6.4cm 0.6cm 6.7cm 0cm,clip]{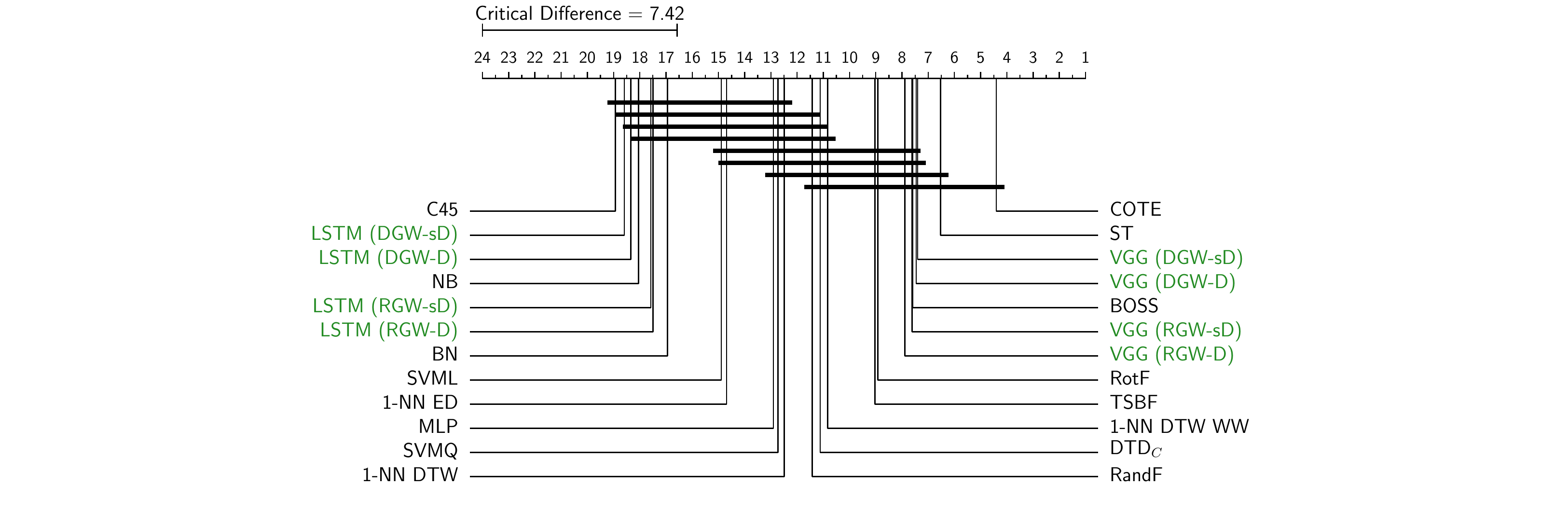}
\caption{Critical difference diagram for the proposed methods (green) and 16 benchmarks.
\vspace{-4mm}
}
\label{fig:cd}
\end{figure}

\subsection{Model Accuracy Comparison}

In Fig.~\ref{fig:cd}, we compare the models with the proposed data augmentation to some of the state-of-the-art methods found in literature. 
In the past, there have been many proposed methods of classifying the 2015 UCR Time Series Archive. 
In order to collect comparisons, we use the best classifier from each category of the great time series classification bake off~\cite{Bagnall_2016} plus all of their baseline methods. 
We also include the classic 1-Nearest Neighbor (1-NN) comparisons from~\cite{UCRArchive}. 
Of the reported comparison methods, only Collection of Transformation Ensembles~(COTE)~\cite{Bagnall_2016_COTE} and Shapelet Transform~(ST)~\cite{Hills_2013} had better results than the proposed augmentation with VGG. 
However, our purpose is not to specifically get the best results overall, but instead to propose a state-of-the-art data augmentation method that can be used with any classifier.

\section{Conclusion}
In this work, we proposed a new data augmentation method for time series based on time warping using DTW. 
We use the time step relationships from the warping path generated by DTW to warp one time series to a second reference time series. 
In doing so, the generated pattern has the local features of the first time series and the time step properties of the second time series. 

Furthermore, we demonstrate that the proposed method can be extended in two ways. 
First, we apply shapeDTW in order to preserve the relationships between neighboring elements in the warping. 
Second, the results are further improved by using the most discriminative sample in a batch as a teacher. 
By using specific discriminative samples as references, we are able to direct the augmentation to target useful teachers.

In the future, we plan on applying the proposed data augmentation method to new applications and new models. 
In addition, we will further explore the properties and characteristics of guided warping. 
An implementation of the proposed method can be found at \url{https://github.com/uchidalab/time_series_augmentation}.


\section*{Acknowledgment}
This research was partially supported by MEXT-Japan (Grant No. J17H06100).


\bibliographystyle{IEEEtran}
\bibliography{icpr}

\end{document}